\documentclass[10pt,twocolumn,letterpaper]{article}

\usepackage[pagenumbers]{cvpr} 

\usepackage{times}
\usepackage{epsfig}
\usepackage{graphicx}
\usepackage{amsmath}
\usepackage{amssymb}
\usepackage{multirow}
\usepackage{graphics}
\usepackage{threeparttable}
\usepackage{color}
\usepackage[normalem]{ulem}
\usepackage{float}
\usepackage{amsfonts}
\usepackage{bm}
\usepackage{bbm}
\usepackage{enumitem}
\usepackage[numbers]{natbib}
\usepackage{subcaption}
\usepackage{array}
\usepackage[table]{xcolor}
\usepackage{colortbl}
\usepackage{soul}
\usepackage[misc]{ifsym}
\usepackage{mathtools}
\usepackage{siunitx}
\usepackage[nopar]{lipsum}
\usepackage{listings}
\usepackage[marginal]{footmisc}

\usepackage{dblfloatfix}
\usepackage{transparent}
\usepackage{arydshln}

\definecolor{myy}{RGB}{126,95,0}
\definecolor{mygray}{gray}{.9}
\definecolor{bblue}{RGB}{30,80,120}
\definecolor{mygray1}{gray}{.7}
\definecolor{ggray}{RGB}{127,127,127}

\newcommand{\pub}[1]{\color{gray}{\tiny{[{#1}]}}}
\newcommand{\baseline}[1]{\color{ggray}{\scriptsize{{#1}}}}

\def\1{\mathbbm{1}}

\usepackage{tabulary}
\newcolumntype{I}{!{\vrule width 1pt}}

\newcolumntype{x}[1]{>{\centering\arraybackslash}p{#1pt}}
\newcolumntype{y}[1]{>{\raggedright\arraybackslash}p{#1pt}}
\newcolumntype{z}[1]{>{\raggedleft\arraybackslash}p{#1pt}}
\newlength\savewidth

\makeatletter
\newcommand{\thickhline}{%
	\noalign {\ifnum 0=`}\fi \hrule height 1pt
	\futurelet \reserved@a \@xhline
}
\makeatother

\newcommand\blfootnote[1]{%
	\begingroup
	\renewcommand\thefootnote{}\footnote{#1}%
	\addtocounter{footnote}{-1}%
	\endgroup
}

\newcommand{\tablestyle}[2]{\setlength{\tabcolsep}{#1}\renewcommand{\arraystretch}{#2}\centering\footnotesize}

\usepackage[pagebackref=true,breaklinks=true,letterpaper=true,colorlinks,bookmarks=false]{hyperref}

\usepackage[capitalize]{cleveref}
\crefname{section}{§}{§§}
\Crefname{section}{§}{§§}

\def\coco{COCO 2014}

\usepackage[accsupp]{axessibility}

\def\cvprPaperID{2450} 
\def\confName{CVPR}
\def\confYear{2022}

\begin{document}

\title{Regional Semantic Contrast and Aggregation for Weakly Supervised Semantic Segmentation}

\author{Tianfei Zhou$^{1,*}$,~~Meijie Zhang$^{2,*}$,~~Fang Zhao$^{3}$,~~Jianwu Li$^{2,\dagger}$\\
	\small{$^1$ Computer Vision Lab, ETH Zurich}~~~\small{$^2$ Beijing Institute of Technology}~~~\small{$^3$
		Inception Institute of AI}\\
	\small\url{https://github.com/maeve07/RCA.git}
}

\maketitle

\begin{abstract}

		\vspace{-.5em}
	Learning semantic segmentation from weakly-labeled (\eg, image tags only) data  is challenging since it is hard to infer dense object regions from sparse semantic tags. Despite being broadly studied, most current efforts directly learn from limited semantic annotations carried by individual image or image pairs, and struggle to obtain integral localization maps. Our work alleviates this from a novel perspective, by exploring rich {semantic contexts} synergistically among {abundant} weakly-labeled training data for network learning \textbf{and} inference. In particular, we propose \underline{r}egional semantic \underline{c}ontrast and \underline{a}ggregation (RCA)~\blfootnote{$^*$ Equal contributions; $^\dagger$ Corresponding author: Jianwu Li.}\!. RCA is equipped with a regional memory bank to store massive, diverse object patterns appearing in training data, which acts as strong support for exploration of dataset-level semantic structure. Particularly, we propose \textbf{i) semantic contrast} to drive network learning by contrasting massive categorical object regions, leading to a more holistic object pattern understanding,
	and \textbf{ii) semantic aggregation} to  gather diverse relational contexts in the memory to enrich semantic representations. In this manner, RCA earns a strong capability of fine-grained semantic understanding, and eventually establishes new state-of-the-art results on two popular benchmarks, \ie, PASCAL VOC 2012 and COCO 2014.
\vspace{-1em}

\end{abstract}

\section{Introduction}\label{sec:intro}

Semantic segmentation continues to be a fundamental task in computer vision, with numerous applications in autonomous driving, robotics, human-computer interactions and medical imaging analysis. While fully supervised systems have achieved tremendous progress, they are limited by the availability of pixel-level annotations, often harvested at great cost, even with smart interfaces~\cite{bearman2016s}. Weakly supervised semantic segmentation (\mbox{WSSS}) alternatively investigates whether this task can be adequately addressed with efficient and weak supervisory signals (\eg, image labels~\cite{wei2018revisiting,jiang2019integral,araslanov2020single,li2021group}, scribbles~\cite{lin2016scribblesup,vernaza2017learning,Liang2022TEL}, bounding boxes~\cite{dai2015boxsup,song2019box,lee2021bbam,oh2021background}). This work studies the form of image-level labels which can be obtained effortlessly,  and thus have been widely embraced in mainstream approaches.

In the absence of the true ``{image label}'' to ``{object region}" correspondence in training data, learning to map visual concepts to pixel regions is particularly challenging. The seminal work, \ie, class activation mapping (\mbox{CAM})~\cite{zhou2016learning}, solves this by mining regions from internal activation of an image classifier. However, the technique is prone to give sparse and incomplete object estimations, since the classifier is only driven to activate a small proportion of features with strong discriminative capability. To address this, {a prevalent of subsequent efforts} strive to learn more complete object regions by, \eg, {region growing} to expand initial responses~\cite{kolesnikov2016seed,wang2018weakly,huang2018weakly}, {adversarial erasing} in a hide-and-seek fashion~\cite{hou2018self,Singh_2017,wei2017object,lee2019ficklenet},  {feature enrichment} to collect within-image  contexts~\cite{wei2018revisiting,yao2021non}, {seeking auxiliary  saliency supervisions}~\cite{zeng2019joint,lee2021railroad,xu2021leveraging}, or self-supervised learning with pre-designed pretext tasks~\cite{shimoda2019self,wang2020self,chang2020weakly}.

\begin{figure}[t]
	\centering
	\includegraphics[width=\columnwidth]{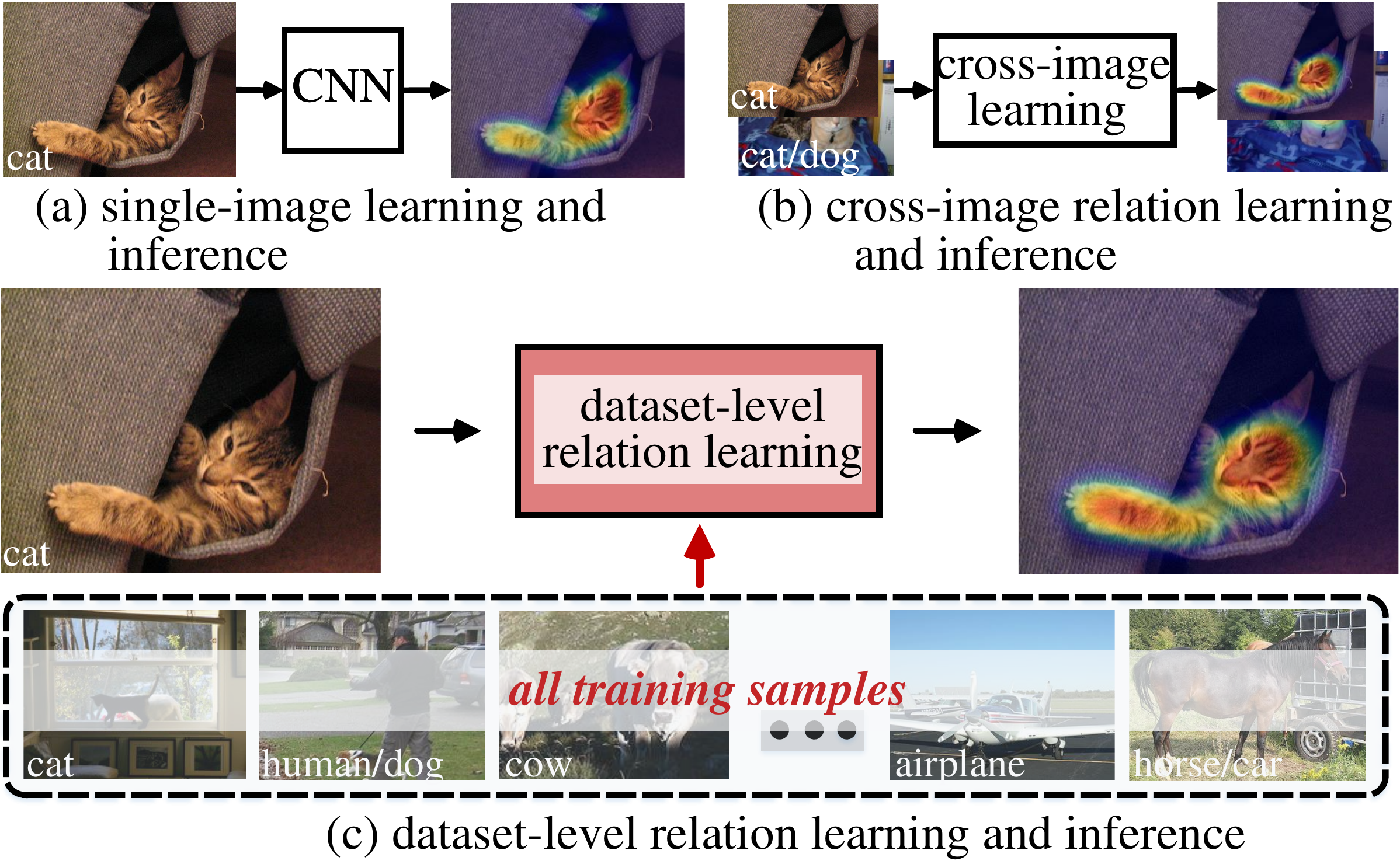}
	\put(-188,101.5){\scriptsize ~(\eg,  \cite{wei2018revisiting,jiang2019integral,araslanov2020single,wang2020weakly,lee2021railroad,xu2021leveraging})}
	\put(-49,101.5){\scriptsize ~(\eg,  \cite{fan2020cian,sun2020mining,li2021group})}
	\vspace{-5pt}
	\captionsetup{font=small}
	\caption{\small \textbf{The main idea} promoted throughout the paper is that semantic contexts subserve localization of individual objects in WSSS. Our RCA thus performs dataset-level relation learning (c) to mine rich contextual knowledge from massive (ideally all) training samples, rather than from an individual image (a) or image pair (b). This enables our model to procure in-depth semantic pattern understanding, improving object localization eventually.}
	\label{fig:motivation}
	\vspace{-10pt}
\end{figure}

Though impressive, these methods use only single-image information for object localization (Fig.$_{\!\!}$~\ref{fig:motivation}$_{\!\!}$~(a)), neglecting inter-image contextual information. Image-level labels not only tell the categories appearing in each individual image, but also  unveil the semantic structure  of all images in the dataset. For each concept (\ie, \texttt{cat} in Fig.~\ref{fig:motivation}), the dataset contains numerous semantically similar but visually different instances; for any two different concepts (\eg, \texttt{cat} and \texttt{dog}), all their instances are semantically different, even though some may look very similar with each other. This \textit{a priori} knowledge should be exploited to gain more accurate semantic pattern understanding.
Though some preliminary attempts~\cite{fan2020cian,zhang2020inter,sun2020mining,zhou2021group} have been made towards this (Fig.$_{\!\!}$~\ref{fig:motivation}$_{\!\!}$~(b)), they focus on  pairwise~\cite{fan2020cian,zhang2020inter,sun2020mining} or quadruplet~\cite{zhou2021group} context modeling in a \textit{limited} number of images, and thus cannot guarantee a sufficient understanding of holistic semantic patterns in the entire dataset. In addition, all these methods favor pixel-wise relation modeling, which is rather difficult due to the lack of proper supervisory signal and causes prohibitive computation cost.

Motivated by above analysis, we propose regional semantic contrast and aggregation (RCA) to maximally exploit  contextual knowledge in visual data (Fig.$_{\!}$~\ref{fig:motivation} (c)), aiming for comprehensive object pattern learning as well as effective CAM inference. In lieu of pixel-level relation modeling in~\cite{fan2020cian,sun2020mining,zhou2021group},  RCA prefers \emph{region-aware} representations that are more efficient and robust to noises. In particular, for each mini-batch image, we divide it into categorical  \emph{pseudo} regions according to an intermediate, coarse CAM, which is learned under the supervision of its single-image label. For each pseudo region, RCA establishes  its relations  to regions in all other images to facilitate dataset-level semantic context learning. For feasible computations, we associate RCA with a continuously-updated memory bank, which collects and preserves meaningful region semantics in the dataset as the training goes, and is applicable to both network learning and inference phases. During training, RCA explores semantic relations of regions in each mini-batch and the memory bank from two novel perspectives:
\begin{itemize}[leftmargin=*]
	\setlength{\itemsep}{0pt}
	\setlength{\parsep}{-2pt}
	\setlength{\parskip}{-0pt}
	\setlength{\leftmargin}{-8pt}
	\vspace{-4pt}
	\item \textit{Semantic contrast}, which lets the model learn to discriminate all possible object regions in the dataset, promoting more holistic object pattern understanding. Particularly, for each pseudo region, {semantic contrast} enforces the network to pull its embedding close to memory embeddings of the same category and push apart those of different. Such a contrastive property well complements the classification objective (for each single image) to improve object representation learning.
	
	\item  \textit{Semantic aggregation}, which allows the model to gather dataset-level  contextual knowledge to yield more meaningful object representations. This is achieved via a non-parametric attention module which  summarizes memory representations  for each  image independently. In comparison with conventional \textit{intra-image} context learning schemes~\cite{chen20182,yuan2020object}, our semantic aggregation focuses on  \textit{inter-image}  context mining, and thus is able to capture more informative dataset-level semantics.

	\vspace{-4pt}
\end{itemize}

These two context modeling schemes  are indispensable to our model. Semantic contrast helps the network to learn more structured object embedding space from a holistic view, while semantic aggregation focuses on  improving  feature representations of each image by collecting diverse semantic contexts. In addition, semantic contrast is essential to maintain unique and informative memory embeddings, which is a prerequisite to yield reliable semantic aggregation. These two components work together to make  RCA a powerful WSSS model (see Table~\ref{table:ablation}).  Our RCA is flexible and can be easily incorporated into existing WSSS models. It shows consistently improved segmentation performance on challenging datasets (\ie, PASCAL VOC 2012~\cite{everingham2010pascal} and COCO 2014~\cite{lin2014microsoft}), on top of state-of-the-art WSSS models (\ie, OAA$^+$~\cite{jiang2019integral}, EPS~\cite{lee2021railroad}).

\noindent\textbf{Main Contributions.} {\textbf{i)}} We study an essential yet long-ignored problem in \mbox{WSSS} to explore rich contexts among weakly labeled training data for network learning. This essentially narrows the gap between image-level semantic concepts and pixel-level  object regions.  Technically, {\textbf{ii)}} we introduce a robust contrastive learning algorithm for semantic contrast, which is able to learn effective representations from imperfect, pseudo region features, as well as   {\textbf{iii)}}  a non-parametric attention model for semantic aggregation to collect rich contextual knowledge from the entire dataset .


\begin{figure*}[!t]
	\centering
	\includegraphics[width=\textwidth]{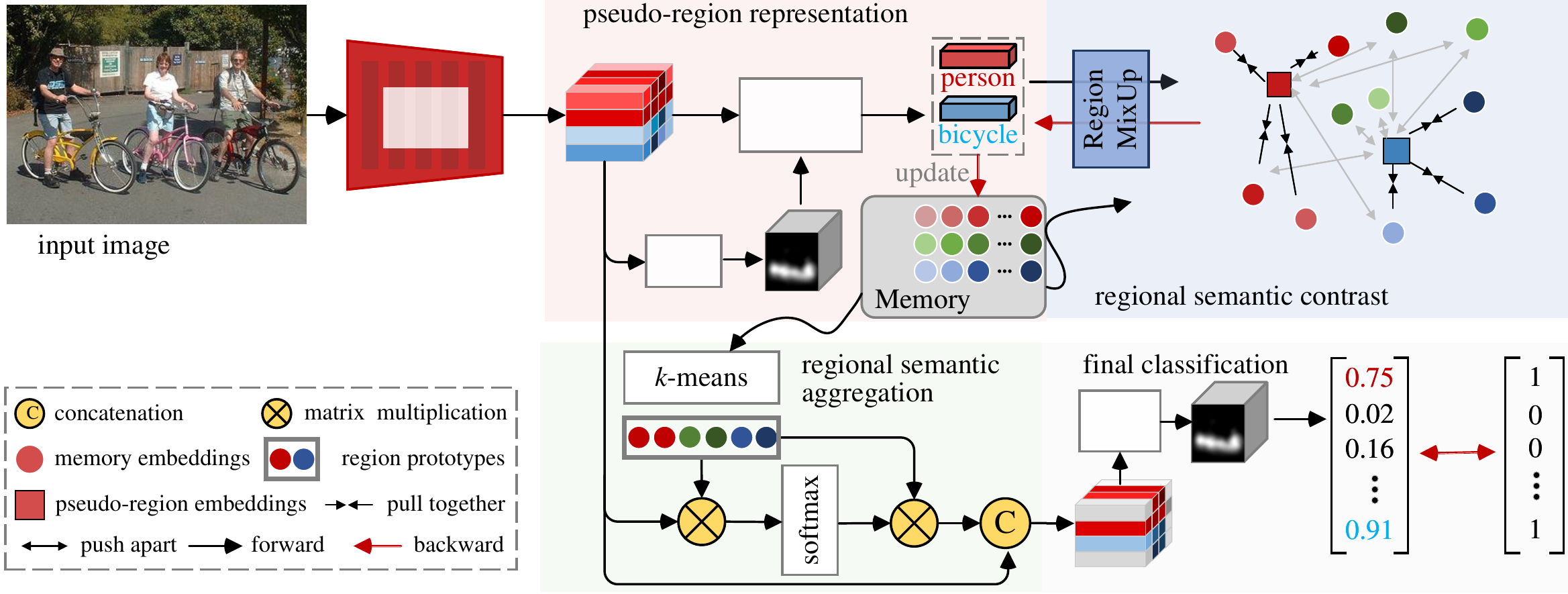}
	\put(-439,106.5){\small$I$}
	\put(-372,148){\small$\mathcal{F}_{\text{FCN}}$}
	\put(-290,102.5){\small$\mathcal{F}_{\text{CAM}}$}
	\put(-152,52.5){\small$\mathcal{F}_{\text{CAM}}$}
	\put(-253,154){\small MAP}
	\put(-255,145){(\small Eq.~\ref{eq:h})}
	\put(-287,137){\small$\bm{F}$}
	\put(-126,10){\small$\hat{\bm{F}}$}
	\put(-260,116){\small$\bm{P}$}
	\put(-102,37){\small$\bm{O}$}
	\put(-300,34){\small$\bm{Q}$}
	\put(-230,12){\small$\bm{S}$}
	\put(-200,12){\small$\bm{F}'$}
	\put(-42,34){\small{\color{red}$\mathcal{L}^{\text{CE}}$}}
	\put(-100,104){\small{\color{red}$\mathcal{L}^{\text{RM-NCE}}$}}
	\put(-187,90){\small$\mathcal{M}$}
	\put(-223,118){\scriptsize$\mathcal{M}_1$}
	\put(-223,109){\scriptsize$\mathcal{M}_2$}
	\put(-223,100){\scriptsize$\mathcal{M}_3$}
	\put(-14,3){\small$\bm{y}$}
	\put(-80,3){\small$\text{GAP}(\bm{O})$}
	\put(-53,90.5){\small(\S\ref{sec:rsc})}
	\put(-199,60){\small(\S\ref{sec:rsa})}
	\put(-205,180){\small(\S\ref{sec:prr})}
	\vspace{-5pt}
	\captionsetup{font=small}
	\caption{\small  Detailed illustration of \textbf{regional semantic contrast and aggregation}. See \S\ref{sec:method} for more details.}
	\label{fig:framework}
	\vspace{-10pt}
\end{figure*}
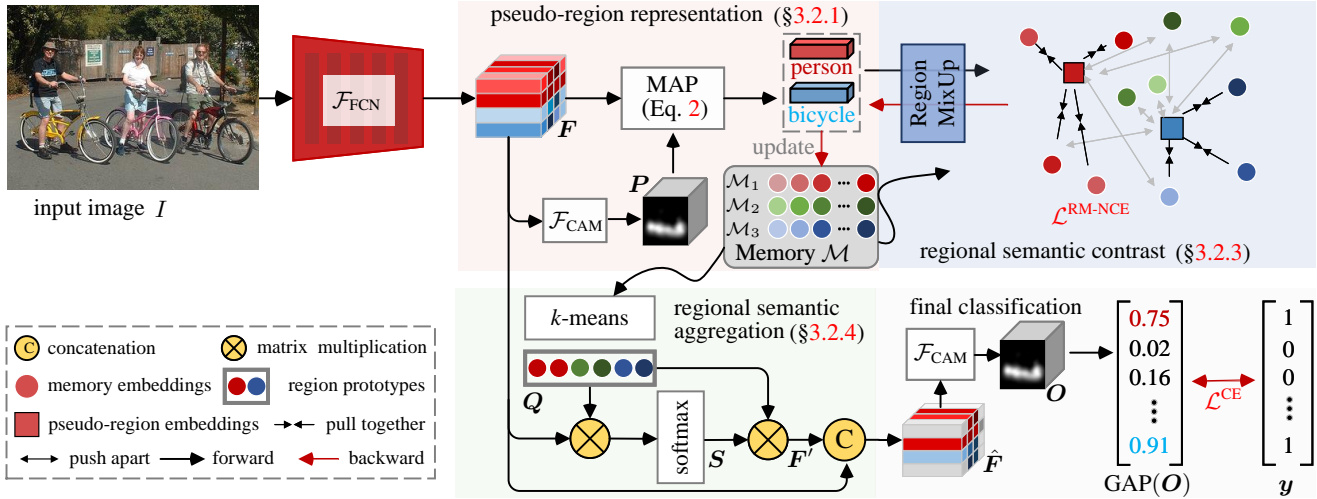

\section{Related Work}\label{sec:relatedwork}

\noindent\textbf{Weakly Supervised Semantic Segmentation} is gaining popularity due to its practical value in reducing the burden of collecting pixel-level annotations at a large scale required by its fully-supervised counterparts \cite{wang2021hierarchical,zhou2021differentiable,zhou2020motion,zhou2021target,wang2021survey}
Here weak supervision may come in diverse forms, \eg, image-level labels~\cite{wei2016stc,chaudhry2017discovering,zhang2020causal,zhou2021group,wang2021multiple}, scribbles~\cite{lin2016scribblesup,vernaza2017learning}, bounding boxes~\cite{dai2015boxsup,khoreva2017simple,song2019box,oh2021background},  point clicks~\cite{bearman2016s,ke2021universal}. Among them, image-level labels gain the most attention due to its minimal annotation demand. However, since only the presence or absence of particular semantics is indicated, the task becomes extremely challenging. The pioneering work of~\cite{zhou2016learning} proposes to obtain coarse object localization maps (\ie, CAMs) from CNN-based image classifiers as seeds to generate pixel-level pseudo segmentation labels. Follow-up works expand coarse CAMs to obtain full extents of object regions by region growing~\cite{kolesnikov2016seed,wei2018revisiting,huang2018weakly}, using stochastic inference~\cite{lee2019ficklenet}, incorporating self-supervised learning~\cite{chang2020weakly, shimoda2019self, wang2020self}, exploring boundary constraints~\cite{chenweakly, lee2021railroad}, or alternatively mining and erasing object regions~\cite{hou2018self, wei2017object, li2018tell}.

Past efforts only consider each image individually, ignoring the rich semantic context across different training images. Recent works~\cite{fan2020cian, sun2020mining} address cross-image semantic mining by computing semantic co-attention between each pair of images, while~\cite{zhou2021group} further enables high-order semantic mining from more images through a graph neural network architecture. Though impressive, these approaches still consider limited semantic context within a \textit{small} number of images (\ie, $2$ in~\cite{fan2020cian, sun2020mining} and $4$ in ~\cite{zhou2021group}). 
In contrast, our approach takes a further step to explore the learning of \emph{rich} relations from a \emph{large} number of weakly annotated data.  It is equipped with a \emph{pseudo-region} memory bank to store region-level semantic embedding for each category, which enables region-aware semantic contrast and aggregation for more comprehensive object pattern mining. 

\noindent\textbf{Contrastive Representation Learning}  is becoming increasingly attractive due to its great potential for \textit{un-/self-supervised} representation learning~\cite{sohn2016improved,oord2018representation,wu2018unsupervised,tian2020contrastive,chen2020simple,he2020momentum}. These approaches learn to compare samples in order to push apart dissimilar (or \emph{negative}) data pairs while pulling together similar (or \emph{positive}) pairs. Some approaches~\cite{caron2018deep,grill2020bootstrap,chen2021exploring} even achieve compelling performance without using any negative pairs. Beyond image-level instance discrimination, recent efforts~\cite{chaitanya2020contrastive,wang2021dense,xie2021propagate} explore pixel- or patch-level discrimination to learn visual representations that generalize better to downstream dense prediction tasks (\eg, semantic segmentation, object detection). Furthermore, \textit{supervised} contrastive learning has been studied in~\cite{khosla2020supervised} for image recognition and in~\cite{wang2021exploring} for supervised semantic segmentation. These methods extend the self-supervised setup (by leveraging label information) to contrast the set of all samples from the same class as positives against {the} negatives from other classes. Inspired by these advances, our approach performs dense contrative learning to improve object localization ability of neural networks, using \textit{weakly supervised} annotations. Our approach is naturally distinguished from the above dense representation learning methods which either neglect any annotations~\cite{chaitanya2020contrastive,wang2021dense,xie2021propagate} or require pixel-level supervisions~\cite{wang2021exploring}.

\noindent\textbf{Relational Context Learning} is popular in image and video segmentation to augment feature embedding of each pixel by gathering useful representations from  its contextual pixels~\cite{zhang2018context,fu2019dual} or regions~\cite{chen20182,yuan2020object}. However, these methods are limited to capturing local contexts within each individual image, ignoring potential semantic contexts across different images. In sharp contrast, our semantic aggregation mines relational semantics across all images of  the entire dataset to gain more informative context learning.

\noindent\textbf{Non-Parametric Memory Bank} has been found feasible to remember a massive number of samples for learning good representations~\cite{wu2018unsupervised,he2020momentum,wang2020cross,wang2021exploring,misra2020self}. Our memory bank is inspired by these efforts, which however, is  unique in that \emph{i)} it stores consistent and expressive region-level semantics inferred from image-level labels; \emph{ii)} more importantly, it is also kept alive in the inference phase to provide holistic contextual knowledge for network inference.

\section{Our Approach}\label{sec:method}

\subsection{Problem Statement}\label{sec:prob}

\noindent\textbf{Task Setup.}  Following the standard setup, each training image ${I}\!\in\!\mathbb{R}^{w_{\!} \times h_{\!} \times 3_{\!}}$ in the dataset $\mathcal{I}$ is associated with only an image-level label vector $\bm{y}\!=\![y_{1}, y_{2},\ldots, y_{L}]\!\in\!\{0,1\}^L$ for $L$ pre-specified  categories. Here, $y_{l}\!=\!1$ indicates the presence of class $l$ in $I$ and $0$ otherwise. Given such coarse annotations, most current solutions follow a  two-phase pipeline to solve the task ``\textit{from classification to segmentation}'', \ie, training a \textit{classification network} first for identifying object regions corresponding to each category, which are then refined to produce pseudo segmentation labels as the supervision of a \textit{semantic segmentation network}. 

\noindent\textbf{Previous Solutions to \mbox{WSSS}.} 
Recent approaches~\cite{jiang2019integral,Zhang_2018,lee2021railroad}, in general, derive class-aware attention maps directly from a fully-convolutional network (\mbox{FCN}), which is proven to produce localization maps with the same quality as \mbox{CAM}~\cite{Zhang_2018}. Particularly, for a mini-batch image ${I}$, its class-aware attention map $\bm{P}$ is generated as follows:
\begin{equation}\small\label{eq:cam}
	\bm{F} = \mathcal{F}_{\text{FCN}}({I})\in\mathbb{R}^{W_{\!} \times_{\!} H_{\!} \times_{\!} D_{\!}}, ~~~
	\bm{P} = \mathcal{F}_{\text{CAM}}(\bm{F})\in\mathbb{R}^{W_{\!} \times_{\!} H_{\!} \times_{\!} L_{\!}}.
\end{equation}
Here, $\mathcal{F}_{\text{FCN}}$ is an FCN network, typically corresponding to the convolutional part of a standard classifier (\eg, VGG \cite{simonyan2014very}, ResNet \cite{he2016deep}).  $\bm{F}$ is the dense  embedding of ${I}$, with $D$ channels and $W_{\!} \times_{\!} H$ spatial size. $\mathcal{F}_{\text{CAM}}$ is a class-aware convolutional layer to produce $\bm{P} \!=\! [\bm{P}_1, \cdots, \bm{P}_L]$, with each map $\bm{P}_l\!\in\!\mathbb{R}^{W_{\!} \times_{\!} H}$ denoting network activation of the $l$-th class. Next, a score vector $\bm{p}\!=\![p_{1},p_{2},\cdots_{\!},p_L]\!\in\!\mathbb{R}^{L}$ is derived from $\bm{P}$ via a global average pooling (GAP) layer, with $p_l\!=\!\text{GAP}(\bm{P}_l)$ being the un-normalized score of the $l$-th class. Finally, $\bm{p}$ is used for multi-label classification.

\noindent\textbf{Our Main Idea.} 
With above  descriptions of existing WSSS solutions, we find that they only exploit limited contextual cues in individual images, causing difficulties for more complete understanding of diverse semantic patterns.  To compensate for this limitation, we introduce a novel method, \ie, RCA, to perform semantic contrast and semantic aggregation over \textit{pseudo regions} of a large number of images (ideally the entire dataset). Both semantic contrast and semantic aggregation are supported by an external pseudo-region memory bank. Next, we will first describe the way to build initial pseudo-region representations (\S\ref{sec:prr}) as well as to construct the memory bank (\S\ref{sec:memorybank}). Then, we elaborate on semantic contrast (\S\ref{sec:rsc}) and semantic aggregation (\S\ref{sec:rsa}). The overall pipeline of RCA is illustrated in Fig.~\ref{fig:framework}.

\subsection{Regional Semantic Contrast and Aggregation}

\subsubsection{Pseudo-Region Representation}\label{sec:prr}

For each mini-batch sample $I$, we convert its dense embedding $\bm{F}$ (Eq.~\ref{eq:cam}) into a set of categorical region representations based on $\bm{P}$ (Eq.~\ref{eq:cam}). Particularly, for the $l$-th category that appears in $I$ (\ie, $y_l\!=\!1$), its region-level semantic information is summarized to a compact embedding vector $\bm{f}_l\!\in\!\mathbb{R}^D$ by masked average pooling (MAP)~\cite{siam2019amp}:
\vspace{-3pt}
\begin{equation}\small\label{eq:h}
	\bm{f}_{l}  =  \frac{\sum_{x=1,y=1}^{W,H}\bm{M}_l(x,y) \bm{F}(x,y)}{\sum_{x=1,y=1}^{W,H}\bm{M}_l(x,y)}~~~\in\mathbb{R}^D,
	\vspace{-3pt}
\end{equation}
where $\bm{M}_{l} \!=\! \1(\bm{P}_{l\!} >_{\!} \mu)\!\in\!\{0,1\}^{W_{\!}\times_{\!}H_{\!}}$ is a binary mask, highlighting only strongly-activated pixels of class $l$ in its activation map (\ie, $\bm{P}_{l}\!\in\!\mathbb{R}^{W_{\!}\times_{\!}H}$). $\1(\cdot)$ is an indicator function, and the threshold $\mu$ is set to the mean value of $\bm{P}_l$. Here $\bm{f}_l$ is compact and lightweight, allowing for  feasible exploration of its relations with a massive number of pseudo regions mining from other samples.

\subsubsection{Pseudo-Region Memory Bank}\label{sec:memorybank}
Taking the inspiration from~\cite{xiao2017joint,wu2018unsupervised}, we setup a non-parametric and dynamic memory bank for RCA to store dataset-level regional semantic information. In particular, the memory bank $\mathcal{M}$ consists of $L$ dictionaries,  \ie, $\mathcal{M}\!=\!\{\mathcal{M}_{1}, \mathcal{M}_{2}, \cdots_{\!}, \mathcal{M}_L\}$, each for one category. Each  entry of $\mathcal{M}_l$ denotes a holistic region-aware representation $\bm{m}_{l}\!\in\!\mathbb{R}^{D}$  of the $l$-th category in image $I$ observed in the whole learning phase. At each training step, the memory bank will be  updated during backward propagation to involve new observations. In particular, the current feature vector $\bm{f}_{l}$ (Eq.~\ref{eq:h}) of image $I$ will be smoothly updated into the memory representation $\bm{m}_l$ as follows:
\vspace{-3pt}
\begin{equation}\small\label{eq:momentum}
	\bm{m}_l \leftarrow \gamma\bm{m}_l + (1-\gamma)\bm{f}_{l},
	\vspace{-1pt}
\end{equation}
where $\gamma$ is the momentum for memory evolution.  We update $\bm{m}_l$ when the $l$-th class appears in $I$ (\ie, $y_l\!=\!1$) and its classification score is higher than a threshold $\nu$, \ie, $p_l\!>\!\nu$. Otherwise, we keep $\bm{m}_l$ as it was.

\noindent\textbf{Memory Mechanism Discussion.} 
Though memory bank has been widely utilized in recent methods~\cite{xiao2017joint,wu2018unsupervised,he2020momentum}, ours shows several unique and appealing characteristics that could lift more advantages to the task of \mbox{WSSS}. \textbf{First}, the memory is compartmentalized enough to compress each potential semantic hypothesis (\ie pseudo-region embedding) in each training sample individually and is able to well encode diverse semantic patterns of each category within weakly-labelled visual data; \textbf{Second}, the momentum updating scheme (Eq.$_{\!}$~\ref{eq:momentum}) not only helps to gain \textit{consistent} memory features for semantic contrast (\S\ref{sec:rsc}) as ~\cite{wu2018unsupervised,he2020momentum}, but more crucially, offers \textit{comprehensive} representations that can accurately describe object semantics. More concretely, Eq.$_{\!}$~\ref{eq:momentum} accumulates all intermediate states (\eg, \{$\bm{f}_l$\}) of each object region produced by the image classifier at different training epochs. These states have shown to be well complementary with each other~\cite{jiang2019integral}, and as a result of Eq.$_{\!}$~\ref{eq:momentum}, each memory feature $\bm{m}_l$ will be gradually promoted to capture a more complete object region as the training goes. This eventually results in informative memory representations after training, which can be leveraged as reliable  contexts for  semantic aggregation (\S\ref{sec:rsa}).

\subsubsection{Regional Semantic Contrast (RSC)} \label{sec:rsc}

We perform semantic contrast over \textit{pseudo}-region semantics for learning more discriminative dense representations. For each categorical pseudo-region embedding $\bm{f}_l$ (Eq.\!~\ref{eq:h}) in image ${I}$, our objective is to increase its similarities to memory features  $\{ \bm{m}_l^+\!\in\!\mathcal{M}_l \}$ of the same class, while reducing the similarities to features  $\{ \bm{m}_l^-\!\in\!\mathcal{M}_{\!}\setminus_{\!}\mathcal{M}_l \}$ of different classes. We achieve this via a region-aware contrastive loss:
\vspace{-3pt}
\begin{equation}\small\label{eq:pNCE}
	\begin{aligned}
	& \mathcal{L}^{\text{NCE}}_{l} (\bm{f}_l, y_l) \\
	& =\frac{1}{|\mathcal{M}_{l}|}\sum_{\bm{m}_{l}^{+\!}\in_{\!}\mathcal{M}_l\!\!} \!\!\!-_{\!}\log\frac{e^{\text{sim}(\bm{f}_{l}, \bm{m}_{l}^{+})/\tau}}{e^{\text{sim}(\bm{f}_{l}, \bm{m}_{l}^{+})/\tau}
		+ \!\!\!\!\sum\limits_{\bm{m}_{l}^{-\!}\in_{\!}\mathcal{M}\setminus_{\!}\mathcal{M}_l\!\!}\!\!\!\!\!\!\!e^{\text{sim}(\bm{f}_{l}, \bm{m}_{l}^{-})/\tau}},\!\!
	\end{aligned}
	\vspace{-3pt}
\end{equation}
where $\tau$ is a temperature hyper-parameter scaling the distribution of distances,$_{\!}$ and$_{\!}$ $\text{sim}(\bm{i}, \bm{j})\!\!=\!\!\frac{\bm{i}\cdot \bm{j}}{\| \bm{i} \|_{2\!}  \| \bm{j} \|_{2\!}}$ is$_{\!}$ the dot product between $\ell_2$-normalized $\bm{i}$ and $\bm{j}$ (\ie, cosine similarity).

Eq.$_{\!}$~\ref{eq:pNCE} falls into the regime of supervised contrastive learning~\cite{khosla2020supervised}, \ie, the labels of {$\bm{f}_l/\bm{m}_l^{+}/\bm{m}_l^{-}$} are given. Differently, in our context, the labels are weak and noisy, posing great challenges  to learn robust representations. 
To alleviate this problem, we develop \emph{region mixup} to regularize Eq.~\ref{eq:pNCE} to learn effective region representations, even from noisy samples. More specifically, for each region $l$ in $I$, we create a mixed region by linearly combining it with a region $l^{-}$ in another mini-batch image. Here we assume that  regions $l$ and $l^-$ are from different categories, \ie, $y_{l}\!\neq\!y_{l^-}$. The embedding of the mixed region  is computed as:
\vspace{-3pt}
\begin{equation}\small\label{eq:mixup}
	\hat{\bm{f}}_l = \omega \bm{f}_l + (1 - \omega) \bm{f}_{l^-},
	\vspace{-1pt}
\end{equation}
where the coefficient $\omega\!\sim\!\mathcal{B}(\beta, \beta)$ follows a Beta distribution $\mathcal{B}(\cdot,\cdot)$ with two shape parameters set to a same $\beta$ \cite{zhang2018mixup}. Then, we define a new region mixup contrastive loss:
\begin{equation}\small\label{eq:mixup-nce}
	\mathcal{L}_l^{\text{RM-NCE}} = \omega \mathcal{L}_l^{\text{NCE}} (\hat{\bm{f}}_l , y_l) + (1  - \omega) \mathcal{L}_l^{\text{NCE}} (\hat{\bm{f}}_l , y_{l^-}).
\end{equation}
It computes two $\mathcal{L}_l^{\text{NCE}}$ losses with respect to $y_l$ and $y_{l^-}$, which are  combined  by the same weight $\omega$ used for region mixup (Eq.~\ref{eq:mixup}). Eq.~\ref{eq:mixup-nce} encourages the network to learn relative similarities for mixed regions, regularizing the model to learn robust representations from label-imperfect samples.

\subsubsection{Regional Semantic Aggregation (RSA)}\label{sec:rsa}

Context is widely recognized to be significant for pixel understanding~\cite{zhang2018context,yuan2020object,jin2021mining}, but prior approaches focus on intra-image context modeling, ignoring rich and valuable inter-image contexts. To alleviate this, we devise semantic aggregation to exploit dataset-level context cues in the memory bank for enhancing semantic understanding.  As stated in \S\ref{sec:memorybank}, our memory bank offers massive  signatures of semantic regions. While a large-scale memory bank could benefit semantic contrast~\cite{he2020momentum}, it contains over-complete (or redundant) representations and some are even noisy, making accurate context learning difficult. In addition, directly aggregating large-scale representations is computationally expensive, and  will greatly slow down the learning and inference procedures. 

To address these problems, we  compress the over-complete memory representations into a compact set of representative prototypes. For each class $l$, we do $k$-means clustering over all features  in $\mathcal{M}_l$ to obtain $K$ prototype vectors (\ie, class centroids), organized in a matrix form  $\bm{Q}_l\!\in\!\mathbb{R}^{K\times D}$.  Here we use multiple prototypes (\ie, $K\!>\!1$) for each class to account for significant intra-class variations.
Next, all the categorical prototypes derived from the memroy bank $\mathcal{M}$ are concatenated together, delivering a holistic prototypical representation $\bm{Q} \!=\! [\bm{Q}_1, \cdots_{\!}, \bm{Q}_L] \in\mathbb{R}^{K_{\!}\times_{\!} D\times_{\!} L_{\!}}$. Then, for each mini-batch image $I$ with feature $\bm{F}\in\mathbb{R}^{W_{\!}\times_{\!} H_{\!}\times_{\!} D}$ (Eq.$_{\!}$~\ref{eq:cam}), {we  calculate its affinity matrix $\bm{S}$ with the prototypical representation $\bm{Q}$ as follows:}
\begin{equation}\small\label{eq:affinity}
	\bm{S} =  \texttt{softmax} (\bm{F}\otimes \bm{Q}^{\top_{\!\!}} )~~~\in\mathbb{R}^{(WH) \times (LK)},
\end{equation}
where $\bm{F}\!\in\!\mathbb{R}^{(WH)\times D}$ and $\bm{Q}\!\in\!\mathbb{R}^{(LK) \times D}$ are flattened into matrix representations for computational convenience. $\otimes$ indicates  matrix multiplication. $\texttt{softmax}(\cdot)$ normalizes each row of the input.  Each entry in $\bm{S}$ reflects the normalized similarity between each row (\ie, feature) in $\bm{F}$ and each column (\ie, prototype) in $\bm{Q}^\top$. Based on the affinity matrix,  the contextual summaries for the feature embedding $\bm{F}$ {w.r.t.}  the prototypical representation $\bm{Q}$ can be computed:
\begin{equation}\small\label{eq:f}
	\bm{F}' = \bm{S}\otimes\bm{Q} ~~~\in\mathbb{R}^{(WH)\times D},
\end{equation}
where ${\bm{F}}'$ denotes an enriched feature representation of $\bm{F}$, which is further reshaped into $\mathbb{R}^{W_{\!}\times_{\!} H_{\!}\times_{\!} D}$. Finally, we concatenate  $\bm{F}'$ and the original feature $\bm{F}$ together:
\vspace{-3pt}
\begin{equation}\small\label{eq:saf}
	\hat{\bm{F}} = [\bm{F}, {\bm{F}}'] ~~~\in \mathbb{R}^{W \times H \times 2D}.
	\vspace{-1pt}
\end{equation}
Here, $\hat{\bm{F}}$  not only encodes intra-image local contexts in $\bm{F}$, but also captures inter-image global  contexts  in $\bm{F}'$, thus enriching the representability for semantic understanding. 

\subsubsection{Class Activation Map Prediction}
Finally, $\hat{\bm{F}}$ is fed into another class-aware convolutional layer $\mathcal{F}_{\text{CAM}}$  (Eq.~\ref{eq:cam}) to produce the final activation maps $\bm{O}$:
\vspace{-3pt}
\begin{equation}\small\label{eq:o}
 \bm{O}=\mathcal{F}_{\text{CAM}}(\hat{\bm{F}})~~\!\in\!\mathbb{R}^{W_{\!}\times_{\!} H_{\!}\times_{\!} L}.
 \vspace{-1pt}
\end{equation}

\subsection{Detailed Network Architecture}\label{sec:arch}

Our classifier is comprised of four major components: \textbf{i)} The {backbone network} $\mathcal{F}_{\text{FCN}}$ (Eq.~\ref{eq:cam}) maps an input image $I$ into a convolutional representation $\bm{F}$. Any FCN network can be used here, and we use two popular ones,  \ie, VGG16~\cite{simonyan2014very} and ResNet38~\cite{he2016deep}, for fair comparison with existing approaches. \textbf{ii)} The {class-wise convolutional layer} $\mathcal{F}_{\text{CAM}}$ (Eq.~\ref{eq:cam})  produces a class-aware attention map from feature embeddings. In our network, two independent $\mathcal{F}_{\text{CAM}}$ are used in Eq.~\ref{eq:cam} and Eq.~\ref{eq:o}, respectively. Each  is implemented as a $1{\times}1$ convolutional layer. \textbf{iii)} The {memory bank} $\mathcal{M}$  stores all region patterns in training data. Note that the memory bank is  removed at the inference phase, with only compressed global prototypical representations kept instead. This reduces the cost to maintain a large memory bank  during model deployment. \textbf{iv)} The loss function of our classifier is as follows:
	\vspace{-3pt}
	\begin{equation}\small\label{eq:loss}
		\begin{aligned}
		\!\!\!\!\!\!\!\mathcal{L} \!=\!\! \sum_{I} &\alpha_1\mathcal{L}^{\text{RM-NCE}\!\!} \!+\! \alpha_2\mathcal{L}^{\text{CE}\!} (\text{GAP}(\bm{P}), \bm{y}) \!+_{\!}\! \mathcal{L}^{\text{CE}\!} (\text{GAP}(\bm{O}), \bm{y}),\!\!\!\!\!\!\!\!\!\!\!\!
		\end{aligned}
			\vspace{-1pt}
	\end{equation}
	where each image $I$ is supervised by the combination of three losses. The first term $\mathcal{L}^{\text{RM-NCE}}$ is the region mixup contrastive loss (Eq.~\ref{eq:mixup-nce}), which is computed as the average loss of all regions appearing in $I$. The second one is {an auxiliary cross-entropy loss} $\mathcal{L}^{\text{CE}}$ for supervising the intermediate CAM prediction $\bm{P}$ (Eq.~\ref{eq:cam}), while the third loss is the main cross-entropy loss  imposing on the final CAM prediction ${\bm{O}}$ (Eq.~\ref{eq:o}).  The coefficients $\alpha_1$ and $\alpha_2$ balance the three  terms.

\section{Experiment}

\subsection{Experimental Setting}\label{sec:settings}

\noindent\textbf{Dataset.} The experiments are conducted on two datasets:\!\!
\begin{itemize}[leftmargin=*]
	\setlength{\itemsep}{0pt}
	\setlength{\parsep}{-2pt}
	\setlength{\parskip}{-0pt}
	\setlength{\leftmargin}{-10pt}
	\vspace{-6pt}
	\item \textbf{PASCAL$_{\!}$ VOC$_{\!}$ 2012$_{\!\!}$}~\cite{everingham2010pascal}$_{\!}$ is a gold standard benchmark for WSSS. It contains $4,\!369$ images, which are split into $1,\!464/1,\!449/1,\!456$ for $\texttt{train}/\texttt{val}/\texttt{test}$, respectively. It provides pixel-level annotations for $21$  categories. As common practices~\cite{huang2018weakly,lee2019ficklenet,zhang2020causal}, we use additional $10,\!582$ images~\cite{hariharan2011semantic} for training.
	
	\item \textbf{\coco}~\cite{lin2014microsoft}  is a more challenging dataset, containing complex contextual interactions of $80$ object classes, which attracts interests to verify the performance of our model in this dataset. We follow the official setting to use $80$K images for \texttt{train} and $40$K images for \texttt{val}. 
	\vspace{-4pt}
\end{itemize}

\definecolor{Gray}{gray}{0.5}
\newcommand{\demph}[1]{\textcolor{Gray}{#1}}
\definecolor{Highlight}{HTML}{39b54a}  
\renewcommand{\hl}[1]{\textcolor{Highlight}{#1}}
\definecolor{mygreen}{HTML}{39b54a}  
\newcommand{\reshl}[2]{
	\textbf{#1} \fontsize{7.5pt}{1em}\selectfont\color{mygreen}{$\uparrow$ \textbf{#2}}
}

\begin{table}[t]
	\centering
	\small
	\tablestyle{1pt}{1.05}
	\begin{tabular}{|x{70}||x{80}|x{70}l|}
		\thickhline
		&  \multicolumn{2}{c}{mIoU (\%)} & \\ \cline{2-4} 
		\multirow{-2}{*}{variant} &  pseudo label  (\texttt{train})  & segmentation (\texttt{val}) & \\ \hline\hline

		\demph{OAA$^+$}& \demph{-} & \demph{65.2}{} &\\ \hline
		
		OAA$^{++}$
		& {68.2} & {67.7} &  \\
		{w/} RSC (\S\ref{sec:rsc})
		& \reshl{69.5}{1.3} & \reshl{69.3}{1.6} & \\ 
		{w/}  RSA (\S\ref{sec:rsa})
		& \reshl{68.5}{0.3} & \reshl{68.6}{0.9} & \\
		{w/}  RSC  \textit{and} RSA & & &\\
		(full model)& \multirow{-2}{*}{\reshl{71.4}{3.2}} & \multirow{-2}{*}{\reshl{70.6}{2.9}} & \\
		\hline
	\end{tabular}
	\vspace{-6pt}
	\captionsetup{font=small}
	\caption{\small\textbf{Ablation study} on  VOC 2012~\cite{everingham2010pascal}.
		``pseudo label'': generated pseudo labels on the \texttt{train} set; ``segmentation'': segmentation results on the \texttt{val} set.
	}
	\vspace{-10pt}
	\label{table:ablation}
	
\end{table}

\noindent\textbf{Evaluation Protocol.} We evaluate RCA in terms of \textbf{i)} semantic segmentation on  VOC 2012 \texttt{val}/\texttt{test} and COCO 2014 \texttt{val}, and \textbf{ii)} quality of generated pseudo segmentation labels on  VOC 2012 \texttt{train}. As conventions~\cite{jiang2019integral,lee2021railroad}, mean intersection-over-union (mIoU) is used as the  metric in both cases. The scores on VOC 2012 \texttt{test} are obtained from the official evaluation server.



\noindent\textbf{Implementation Details.} As stated in~\S\ref{sec:arch}, we test two commonly used backbones (\ie, VGG16 \cite{simonyan2014very}, ResNet38 \cite{he2016deep}) in RCA for the experiments. The weights of the backbones are loaded from ImageNet pre-trained weights. RCA is trained using the SGD optimizer with batch size {$8$}, momentum {$0.9$} and weight decay 5e-4. The initial learning rates are set to 1e-3 for the backbone and 1e-2 for other components, which are reduced by $0.1$ per five epochs. We warm up the network in the first epoch by  using the cross-entropy losses only in Eq.~\ref{eq:loss}, \ie, $\alpha_1=0$. The network is trained  for $30$ epochs in total.  For VOC 2012, we use an adaptive memory size for each class to store all region embeddings in the dataset, while for COCO 2014, the per-class memory size is set to $500$ to avoid significant memory consumption. The $k$-means prototype clustering in \S\ref{sec:rsa} is performed only once at the beginning of each epoch, and the per-class  prototype number is set to $K\!=\!10$ by default. For the hyper-parameters, we empirically set the threshold $\nu$,  momentum $\gamma$, shape parameter $\beta$,  weights $\alpha_1$ and $\alpha_2$ to {$0.7$}, {$0.99$}, $8$, $0.01$ and {$0.4$}, respectively. 

Once the classifier is well trained, we generate class-aware attention maps $\bm{O}$ (Eq.~\ref{eq:o}) for each training image and regard them as foreground seeds. In line with~\cite{wu2021embedded,xu2021leveraging,lee2021railroad,jiang2019integral,li2021group}, we also compute a saliency map for each image using off-the-shelf models to estimate background cues. The final pseudo labels are obtained by combining the foreground and background cues together~\cite{jiang2019integral,li2021group}. Finally, with the pseudo masks as the supervision, we train  DeepLabV2~\cite{chen2017deeplab} using  the default hyper-parameter setting in~\cite{chen2017deeplab}. Dense CRF \cite{krahenbuhl2011efficient} is used as a post-processing routine to refine segmentation boundaries, as in \cite{lee2021railroad,wu2021embedded,li2021pseudo,xu2021leveraging,zhang2021complementary}.

 

\noindent\textbf{Baselines.} 
RCA is flexible and can be easily incorporated into most  WSSS models. In the experiments, we evaluate RCA based on two baselines, \ie, OAA$^{+}$~\cite{jiang2019integral} (due to its popularity) and EPS~\cite{lee2021railroad} (due to its overall best performance). For the conventional OAA$^{+}$, we build a stronger baseline OAA$^{++}$, by replacing its  saliency model with~\cite{liu2019simple}, which is widely-used by recent approaches~\cite{wu2021embedded,xu2021leveraging}.  EPS is the leading WSSS model nowadays; we use it to validate the efficacy of RCA, even with a strong baseline.

\noindent\textbf{Reproducibility.} Our network is implemented in PyTorch and trained on four \mbox{NVIDIA} V100 cards. Testing is conducted on a single \mbox{NVIDIA} \mbox{RTX2080Ti} card. 

\begin{figure}[t]
	\centering
	\includegraphics[width=\columnwidth]{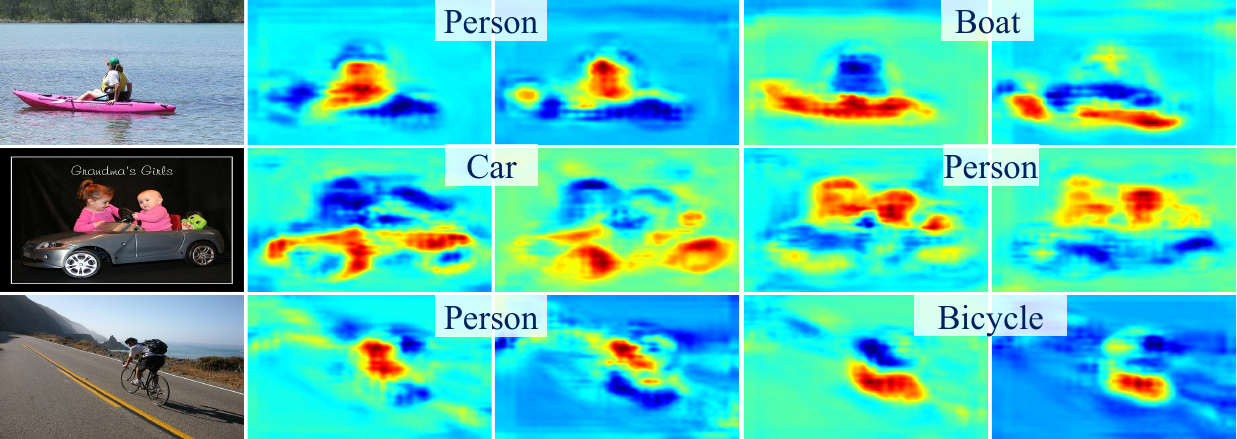}
	\vspace{-18pt}
	\captionsetup{font=small}
	\caption{\small \textbf{Visualization of affinity $\bm{S}$ (Eq.~\ref{eq:affinity})}. Each heatmap correspondes to a column in matrix $\bm{S}$, which is a dot-product between  a particular prototype with image feature $\bm{F}$. See \S\ref{sec:ablation} for details.}
	\label{fig:affinity}
	\vspace{-5pt}
\end{figure}

\subsection{Diagnostic Experiment}\label{sec:ablation}

We first ablate the core designs of RCA in terms of pseudo label quality on  {VOC} 2012 \texttt{train}. \mbox{VGG16} is used as the classification backbone by default.

\noindent\textbf{Semantic Contrast and Semantic Aggregation.} We  investigate the necessity to learn dataset-level visual contexts for WSSS. Table~\ref{table:ablation} summarizes the results. \textbf{First}, the variant ``{w/} RSC'' significantly improves against OAA$^{++}$ in both pseudo label (\ie, $\textbf{1.3\%}$) and segmentation (\ie, $\textbf{1.6\%}$) performance, proving that by contrasting massive object regions, our model fulfills the goal of  more comprehensive object pattern understanding. \textbf{Second}, ``{w/} RSA'' only achieves marginal  performance gains. However, when integrating it with RSC together, our full model (\ie, ``{w/} RSC \textit{and} RSA'') achieves remarkable improvements in comparison with ``{w/} RSC'' ($69.5\%$ vs $\textbf{71.4\%}$ for pseudo label, $69.3\%$ vs $\textbf{70.6\%}$ for segmentation). This reveals that RSC, which helps to obtain informative memory representations, is essential for RSA to perform  reliable context aggregation.

To gain more insights into RSA, we visualize feature-prototype affinity $\bm{S}$ (Eq.~\ref{eq:affinity}) in Fig.~\ref{fig:affinity}. We see that our prototypes are able to attend to semantically meaningful regions, which could benefit object localization.



\noindent\textbf{Region Mixup.} The following table ablates the design of region mixup in \S\ref{sec:rsc}:
\vspace{-6pt}
\begin{table}[H]
	\small
	\centering	
		\tablestyle{1pt}{1.05}
		\begin{tabular}{|z{40}|x{90}x{90}|}
			\thickhline
			variant 
			& {w/o} region mixup (Eq.~\ref{eq:pNCE}) & {w/} region mixup (Eq.~\ref{eq:mixup-nce})  \\ \hline		 
			mIoU (\%)           						 
			& 70.6 & 71.4  \\ \hline		 
	\end{tabular}
	\vspace{-10pt}
\end{table}
\noindent We find that after dropping region mixup, the mIoU score reduces by $0.8\%$. This result reveals that region mixup  indeed helps the model learn more robust representations from noisy data (\ie, pseudo regions), leading to more accurate semantic understanding.

\noindent\textbf{Memory Updating Coefficient $\gamma$.} The table below shows accuracy of generated pseudo segmentation labels with different updating coefficients (Eq.~\ref{eq:momentum}):
\vspace{-6pt}
\begin{table}[H]
	\small
	\centering	
		\tablestyle{7pt}{1.05}
		\begin{tabular}{|r|cccccc|}
			\thickhline
			coefficient $\gamma$                                   
			& 0  & 0.5 & 0.8 & 0.9 & 0.99 & 0.999\\ \hline		 
			mIoU (\%)           						 
			& 69.9 & 70.9 & 71.2 & 71.2 & 71.4 & 70.9 \\ \hline		 
	\end{tabular}
	\vspace{-10pt}
\end{table}
\noindent The optimal value is $\gamma\!=\!0.99$ (our default). Moreover, RCA is robust when $\gamma$ is in $0.8\!\sim\!0.99$, showing that it is beneficial to update the memory in a relatively slow speed, but not too slow (\ie, $\gamma\!=\!0.999$). When $\gamma$ is too small, the performance degrades; at the extreme of \emph{no momentum} (\ie, $\gamma\!=\!0$), the model significantly  degrades. These results support our discussions in \S\ref{sec:memorybank} that momentum updating helps to earn more consistent and comprehensive memory representations, providing powerful assistance for both semantic contrast and semantic aggregation.

\noindent\textbf{Prototype Number $K$.} The following table ablates the role of prototype number $K$ in semantic aggregation (\S\ref{sec:rsa}):
\vspace{-6pt}
\begin{table}[H]
	\small\centering
		\tablestyle{8pt}{1.05}
		\begin{tabular}{|r|cccccc|}
			\thickhline
			$K$                                   & 1  & 10 & 20 & 50 & 100 & all \\ \hline		 
			mIoU (\%)           			   & 70.4 & 71.4 & 71.1 & 71.1 & 71.3 & 70.0  \\ \hline		 	
	\end{tabular}
	\vspace{-10pt}
\end{table}
\noindent  Note that for $K\!=\!1$, we  average all the embeddings in each dictionary to obtain a single prototype vector for each category; for the setting ``all'', we use all memory embeddings as the prototypes without clustering. As seen from the table, RCA shows stable performance when $K$ is in $10\!\sim\!100$. At extreme cases, the model degrades due to severe information loss ($K\!=\!1$) or too many noisy embeddings (``all'').


\noindent\textbf{Memory Size.} By default, our memory bank stores all pseudo regions in the  dataset. However, the following table shows that our model is not sensitive to this setting:
\vspace{-6pt}
\begin{table}[H]
	\small
	\centering	
	\tablestyle{1pt}{1.05}
	\begin{tabular}{|z{50}|x{55}x{55}x{55}|}
		\thickhline
		memory size 
		& 100 & 500 & all  \\ \hline		 
		mIoU (\%)           						 
		& 70.8 & 71.2 & 71.4  \\ \hline		 
	\end{tabular}
	\vspace{-10pt}
\end{table}
\noindent By storing only $100$ or $500$ region embeddings per class, the performance only degrades very slightly. This reveals that our model is scalable to larger-scale datasets (\eg, COCO 2014), for which we cannot afford caching all embeddings.


\subsection{Comparison with Prior Art}

\noindent\textbf{Object Localization.} Table~\ref{table:cam} reports the results of generated pseudo segmentation labels on  VOC 2012 \texttt{train}. Notably, RCA improves OAA$^{++}$ by \textbf{3.2\%} and {\textbf{3.8\%}} when using VGG16 and ResNet38 as the classifier backbones, respectively. It also yields a solid improvement against EPS ($71.4\%$ vs $\textbf{74.1\%}$). These results confirm the strong localization capability of our approach.


\noindent\textbf{Semantic Segmentation.} Table~\ref{table:voc} provides the comparison of RCA against representative methods on VOC 2012 \texttt{val} and \texttt{test}. As seen, RCA  brings solid gains over the two baselines (\ie, OAA$^{++}$ and EPS). Using VGG16 (or ResNet38) as the classification backbone, RCA improves OAA$^{++}$ by {$\textbf{2.9\%}$} ({$\textbf{3.0\%}$}) on \texttt{val},  {$\textbf{3.6\%}$} ({$\textbf{3.4\%}$}) on \texttt{test}. Consistent improvements ({$\textbf{1.3\%}/\textbf{2.0\%}$}) are also seen  for EPS. In addition, RCA{\baseline{+EPS}} sets a new state-of-the-art.


Table~\ref{table:coco} summarizes the segmentation results on  {COCO 2014}~\cite{lin2014microsoft}. We observe that RCA surpasses  OAA$^{+}$ and EPS  by {$\textbf{2.1\%}$}  and {$\textbf{1.1\%}$}, respectively. Remarkably, RCA{\baseline{+EPS}}, which employs VGG16 as the backbone, outperforms many ResNet-based models (\eg, AuxSegNet \cite{xu2021leveraging}).

\begin{figure}[t]
	\centering
	\includegraphics[width=\columnwidth]{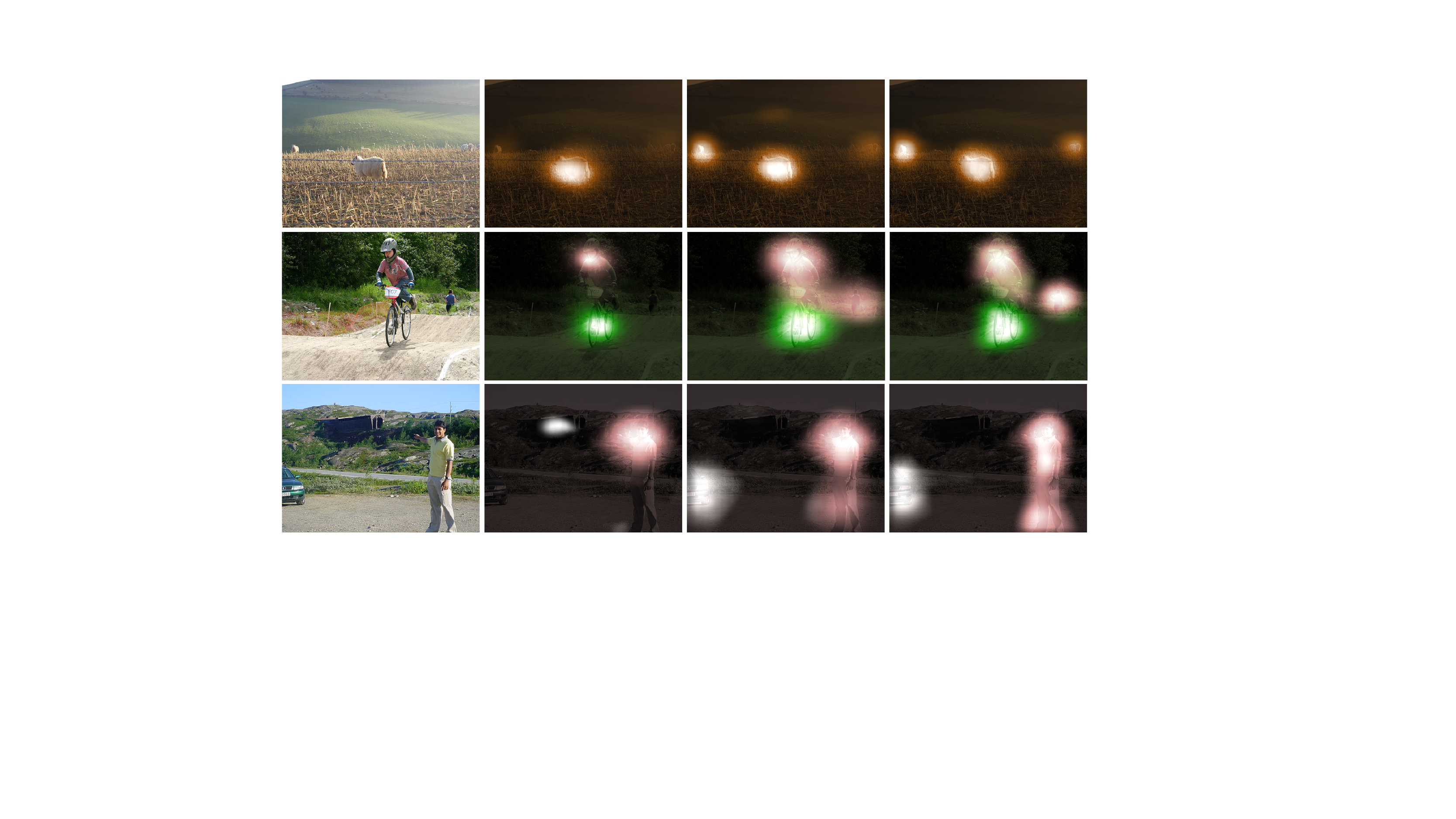}
	\vspace{-18pt}
	\captionsetup{font=small}
	\caption{\small\textbf{Visualization of class activation maps} on VOC 2012 \texttt{train}. From left to right: input images, results of OAA$^{++}$, results from $\bm{P}$ (Eq.~\ref{eq:cam}) and $\bm{O}$ (Eq.~\ref{eq:o}) of our full model.}
	\label{fig:cam}
	\vspace{-6pt}
\end{figure}

\definecolor{Gray}{gray}{0.5}
\newcommand{\std}[1]{{\fontsize{5pt}{1em}\selectfont ~~$_\pm$$_{\text{#1}}$}}
\newcommand{\res}[3]{
	\tablestyle{1pt}{1}
	\begin{tabular}{z{16}y{18}}
		{#1} &
		\fontsize{7.5pt}{1em}\selectfont{~(${#2}${{#3}})}
\end{tabular}}

\begin{table}[t]
	\small
	
	\centering
	\tablestyle{1pt}{1.0}
	\begin{tabular}{|z{90}|x{65}||x{65}|}
		\thickhline
		method &
		backbone &
		mIoU (\%)  \\ 
		\hline\hline
		{SS-WSSS~\pub{CVPR20}{~\cite{araslanov2020single}}}
		& {ResNet38} &  {62.2}   \\
		
		{ICD~\pub{CVPR20}~\cite{fan2020learning}}        
		& {VGG16} &  {62.2}  \\
		
		{SubCat~\pub{CVPR20}{~\cite{chang2020weakly}}}   
		& ResNet38 &  {63.4} \\
		
		{CONTA~\pub{NeurIPS20}{~\cite{zhang2020causal}}}   
		& ResNet38 &  {65.4} \\
		
		{GroupWSSS~\pub{TIP21}{~\cite{zhou2021group}}}   
		& VGG16 &  {65.7}\\
		
		{IRNet~\pub{CVPR19}{~\cite{ahn2019weakly}}}    
		& ResNet50 & {66.5} \\ 
		
		{BES~\pub{ECCV20}{~\cite{chenweakly}}}    
		& ResNet50 &  {67.2}\\
		
		{EDAM~\pub{CVPR21}{~\cite{wu2021embedded}}}   
		& ResNet38 &  {68.1}\\ \hline
		
		
		OAA$^{++}$
		&\multirow{2}{*}{VGG16}  & {68.2}\\ 
		
		{\textbf{RCA}}\baseline{+OAA$^{++}$}
		&  & \reshl{\textbf{71.4}}{3.2} \\ \hdashline

		OAA$^{++}$
		& \multirow{2}{*}{ResNet38}  & {69.4}\\ 
		
		{\textbf{RCA}}\baseline{+OAA$^{++}$}
		&  & \reshl{\textbf{73.2}}{3.8}  \\ \hline
		
		%
		
		{EPS~\pub{CVPR21}{~\cite{lee2021railroad}}}  
		& \multirow{2}{*}{ResNet38} & {71.4}\\ 
		
		{\textbf{RCA}}\baseline{+EPS}
		&  & \reshl{\textbf{74.1}}{2.7} \\ \hline
		
	\end{tabular}
	\vspace{-6pt}
	\captionsetup{font=small}
	\caption{\small\textbf{Quantitative performance of pseudo segmentation labels} on  VOC 2012~\cite{everingham2010pascal} \texttt{train}.}
	\vspace{-10pt}
	\label{table:cam}
\end{table}


\begin{figure*}[t]
	\centering
	\includegraphics[width=\textwidth]{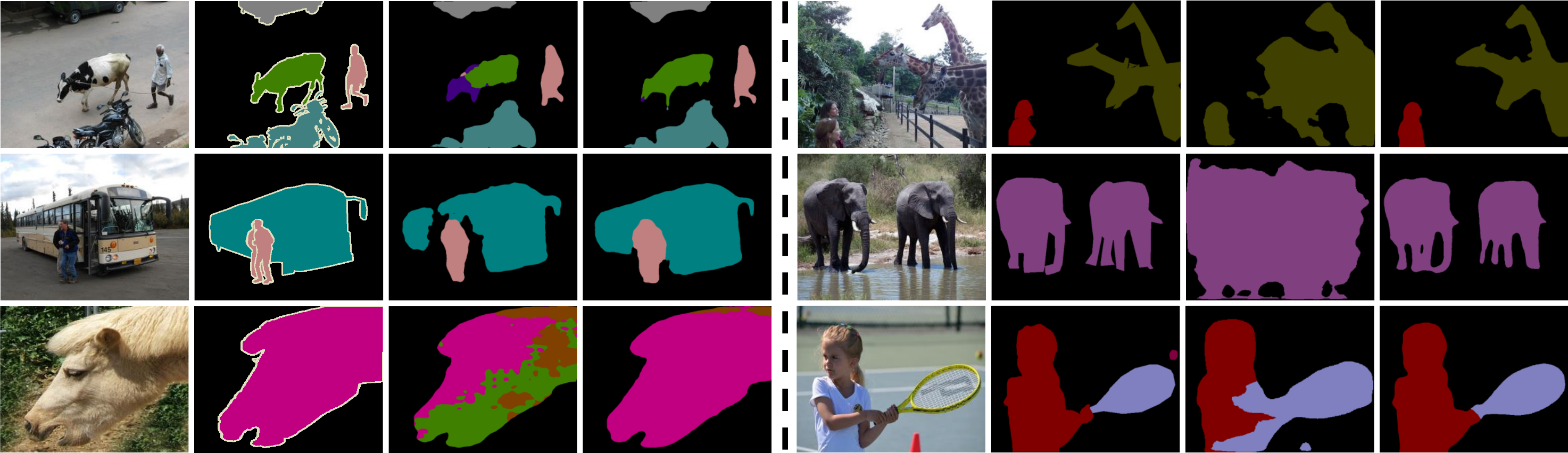}
	\vspace{-18pt}
	\captionsetup{font=small}
	\caption{\small\textbf{Qualitative segmentation results} on VOC 2012 \texttt{val} (left) and COCO 2014 \texttt{val} (right). From left to right: input images, ground-truths, segmentation results of OAA$^{++}$ as well as our RCA.}
	\label{fig:vis}
	\vspace{-8pt}
\end{figure*}


\begin{table}[t]
	\small
	\centering
	
	\tablestyle{1pt}{1.0}
	\begin{tabular}{|z{90}||x{65}x{65}|}
		\thickhline
		&  \multicolumn{2}{c|}{mIoU (\%)} \\ 
		\multirow{-2}{*}{method} &  \texttt{val} & \texttt{test} \\ \hline\hline

		
		$^\ddagger$SSNet~\pub{ICCV19}{~\cite{zeng2019joint}}  
		& {63.3}& {64.3} \\
		$^\ddagger$RNet~\pub{CVPR19}{~\cite{ahn2019weakly}} 
		& {63.5} & {64.8} \\ 
		$^*$CIAN~\pub{AAAI20}{~\cite{fan2020cian}}    
		& {64.3} & {65.3} \\
		$^*$FickleNet~\pub{CVPR19}{~\cite{lee2019ficklenet}}   
		& {64.9} & {65.3} \\
		$^\dagger$SSDD~\pub{ICCV19}{~\cite{shimoda2019self}}    
		& {64.9}& {65.5} \\
		$^\dagger$SEAM~\pub{CVPR20}{~\cite{wang2020self}}    
		& {64.5} & {65.7} \\
		
		$^\dagger$SubCat~\pub{CVPR20}{~\cite{chang2020weakly}}    
		& {66.1} & {65.9} \\
		$^*$OAA$^+$~\pub{ICCV19}{~\cite{jiang2019integral}}   
		& {65.2} & {66.4} \\
		$^\ddagger$BES~\pub{ECCV20}{~\cite{chenweakly}}   
		& {65.7} & {66.6} \\
		$^\dagger$CONTA~\pub{NeurIPS20}{~\cite{zhang2020causal}}    
		& {66.1} & {66.7} \\
		$^*$MCIS~\pub{ECCV20}{~\cite{sun2020mining}}   
		& {66.2} & {66.9}\\
		
		$^*$ICD~\pub{CVPR20}{~\cite{fan2020learning}}  
		& {67.8} & {68.0} \\
		$^\dagger$CPN~\pub{ICCV21}{~\cite{zhang2021complementary}}  
		& {67.8} & {68.5} \\

		$^*$NSROM~\pub{CVPR21}{~\cite{yao2021non}}   
		& {68.3} & {68.5} \\
				$^\dagger$AuxSegNet~\pub{ICCV21}{~\cite{xu2021leveraging}}  
		& {69.0} & {68.6} \\
		$^\ddagger$PMM~\pub{ICCV21}~\cite{li2021pseudo} 
		& {68.5} & {69.0} \\

				$^*$GroupWSSS~\pub{TIP21}{~\cite{zhou2021group}}  
		& {68.7} & {69.0} \\
		$^\dagger$EDAM~\pub{CVPR21}{~\cite{wu2021embedded}}   
		& {70.9} & {70.6} \\ 
		
		$^\dagger$SPML~\pub{ICLR21}~\cite{ke2021universal}   & {69.5} & {71.6} \\ \hline

		$^*$OAA$^{++}$
		& {67.7}& {67.4} \\
		
		$^*${\textbf{RCA}}\baseline{+OAA$^{++}$}
		& \reshl{{70.6}}{2.9} & \reshl{{71.0}}{3.6} \\  \hdashline
		
		$^\dagger$OAA$^{++}$
		& {68.1} & {68.2} \\
		
		$^\dagger${\textbf{RCA}}\baseline{+OAA$^{++}$}
		& \reshl{{71.1}}{3.0} & \reshl{{71.6}}{3.4}\\  
		
		\hline
		
		$^\dagger$EPS~\pub{CVPR21}{~\cite{lee2021railroad}}   
		& {70.9}{} & {70.8}\\ 
		
		$^\dagger${\textbf{RCA}}\baseline{+EPS~\pub{CVPR21}{~\cite{lee2021railroad}}}
		& \reshl{{72.2}}{1.3} & \reshl{{72.8}}{2.0}\\  
		\hline
	\end{tabular}
	\vspace{-6pt}
	\captionsetup{font=small}
	\caption{\small\textbf{Quantitative  performance} on VOC 2012~\cite{everingham2010pascal} \texttt{val} and \texttt{test}. All models use ResNet as the segmentation backbone. $^*$, $^\dagger$ and $^\ddagger$ denote models using VGG16, ResNet38 or ResNet50 as the classification  backbone, respectively.}
	\vspace{-10pt}
	\label{table:voc}
\end{table}

\begin{table}[t]
	\small
	\tablestyle{1pt}{1.0}
	\begin{tabular}{|z{90}|x{65}||x{65}|}
		\thickhline
		{method} & {backbone}  &  mIoU (\%) \\ \hline\hline
		BFBP~\pub{ECCV16}{~\cite{saleh2016built}} 		 		
		& VGG16 & {20.4}{} \\ 
		SEC~\pub{ECCV16}{~\cite{kolesnikov2016seed}}  		
		& VGG16 & {22.4}{} \\  	
		DSRG~\pub{CVPR18}{~\cite{huang2018weakly}}    	
		& VGG16 & {26.0}{} \\ 
		IAL~\pub{IJCV20}{~\cite{wang2020weakly}}    		
		& VGG16 & {27.7}{} \\ 
		GroupWSSS~\pub{TIP21}~\cite{zhou2021group} 			
		& VGG16 & {28.7}{} \\ 
		ADL~\pub{PAMI20}~\cite{choe2020attention} 
		& VGG16 & {30.8}{} \\ 
		SEAM~\pub{CVPR20}~\cite{wang2020self}  				
		& ResNet38 & {32.8}{ }\\ 
		CONTA~\pub{NeurIPS20}~\cite{zhang2020causal}  
		& ResNet38 & {32.8}{} \\ 
		AuxSegNet~\pub{ICCV21}~\cite{xu2021leveraging} 
		& ResNet38 & {33.9}{} \\  \hline
		OAA$^{+}$~\pub{ICCV19}{~\cite{jiang2019integral}} 
		& VGG16 & {24.6}{} \\ 
		\textbf{RCA}\baseline{+OAA$^{+}$~\pub{ICCV19}{~\cite{jiang2019integral}}}
		& VGG16 & \reshl{\textbf{26.7}}{2.1} \\ \hline
		EPS~\pub{CVPR21}{~\cite{lee2021railroad}}
		& VGG16 & {35.7}{} \\ 
		\textbf{RCA}\baseline{+EPS~\pub{CVPR21}{~\cite{lee2021railroad}}}
		& VGG16 & \reshl{\textbf{36.8}}{1.1} \\ 
		\hline

	\end{tabular}
	\vspace{-6pt}
	\captionsetup{font=small}
	\caption{\small\textbf{Quantitative performance} on COCO 2014~\cite{lin2014microsoft} \texttt{val}. 	
	}
	\vspace{-12pt}
	\label{table:coco}
\end{table}

\subsection{Visualization Result} 

\noindent\textbf{Object Localization.} 
Fig.~\ref{fig:cam} depicts some representative CAM predictions of OAA$^{++}$ and RCA for training samples in PASCAL VOC 2012. As observed, our RCA is able to produce more integral object localization results across various challenging situations (\eg, tiny objects, scale variations). In addition, the final CAM predictions (Eq.~\ref{eq:o}) {are} more accurate than the intermediate ones (Eq.~\ref{eq:cam}), demonstrating the effectiveness of our core designs.

\noindent\textbf{Semantic Segmentation.} Fig.$_{\!}$~\ref{fig:vis} illustrates some qualitative segmentation results of OAA$^{++}$ and  RCA on  VOC 2012 \texttt{val} and COCO 2014 \texttt{val}. We find that RCA achieves more accurate segmentation results than OAA$^{++}$, showing remarkable capabilities in handling complex scenes, such as  small/large objects, multiple instances, occlusions.

\section{Conclusion}

In this work, we present a novel approach, RCA, to learn semantic segmentation using image-level supervision only. To alleviate the limited available knowledge carried by image labels, our approach explores the possibility to discover rich semantic contexts from weakly-labeled training data for  learning. In particular, RCA is equipped with a continuously updated memory bank for storing massive historical pseudo-region features. The semantic relations between memory contents and mini-batch training samples are sufficiently exploited as additional supervisory signals (by semantic contrast) or holistic contextual cues (by semantic aggregation) to improve network learning and inference. Our approach is effective and principled, with extensive experiments manifesting its leading performance on popular benchmarks, \ie, PASCAL VOC 2012 and COCO 2014.


{
\small
\bibliographystyle{ieee_fullname}
\bibliography{egbib}

\begin{thebibliography}{10}\itemsep=-1pt

\bibitem{ahn2019weakly}
Jiwoon Ahn, Sunghyun Cho, and Suha Kwak.
\newblock Weakly supervised learning of instance segmentation with inter-pixel
  relations.
\newblock In {\em CVPR}, 2019.

\bibitem{araslanov2020single}
Nikita Araslanov and Stefan Roth.
\newblock Single-stage semantic segmentation from image labels.
\newblock In {\em CVPR}, 2020.

\bibitem{bearman2016s}
Amy Bearman, Olga Russakovsky, Vittorio Ferrari, and Li Fei-Fei.
\newblock What’s the point: Semantic segmentation with point supervision.
\newblock In {\em ECCV}, 2016.

\bibitem{caron2018deep}
Mathilde Caron, Piotr Bojanowski, Armand Joulin, and Matthijs Douze.
\newblock Deep clustering for unsupervised learning of visual features.
\newblock In {\em ECCV}, 2018.

\bibitem{chaitanya2020contrastive}
Krishna Chaitanya, Ertunc Erdil, Neerav Karani, and Ender Konukoglu.
\newblock Contrastive learning of global and local features for medical image
  segmentation with limited annotations.
\newblock In {\em NeurIPS}, 2020.

\bibitem{chang2020weakly}
Yu-Ting Chang, Qiaosong Wang, Wei-Chih Hung, Robinson Piramuthu, Yi-Hsuan Tsai,
  and Ming-Hsuan Yang.
\newblock Weakly-supervised semantic segmentation via sub-category exploration.
\newblock In {\em CVPR}, 2020.

\bibitem{chaudhry2017discovering}
Arslan Chaudhry, Puneet~K Dokania, and Philip~HS Torr.
\newblock Discovering class-specific pixels for weakly-supervised semantic
  segmentation.
\newblock {\em arXiv preprint arXiv:1707.05821}, 2017.

\bibitem{chenweakly}
Liyi Chen, Weiwei Wu, Chenchen Fu, Xiao Han, and Yuntao Zhang.
\newblock Weakly supervised semantic segmentation with boundary exploration.
\newblock In {\em ECCV}, 2020.

\bibitem{chen2017deeplab}
Liang-Chieh Chen, George Papandreou, Iasonas Kokkinos, Kevin Murphy, and Alan~L
  Yuille.
\newblock Deeplab: Semantic image segmentation with deep convolutional nets,
  atrous convolution, and fully connected crfs.
\newblock {\em IEEE TPAMI}, 40(4):834--848, 2017.

\bibitem{chen2020simple}
Ting Chen, Simon Kornblith, Mohammad Norouzi, and Geoffrey Hinton.
\newblock A simple framework for contrastive learning of visual
  representations.
\newblock 2020.

\bibitem{chen2021exploring}
Xinlei Chen and Kaiming He.
\newblock Exploring simple siamese representation learning.
\newblock In {\em CVPR}, 2021.

\bibitem{chen20182}
Yunpeng Chen, Yannis Kalantidis, Jianshu Li, Shuicheng Yan, and Jiashi Feng.
\newblock A2-nets: Double attention networks.
\newblock In {\em NeurIPS}, 2018.

\bibitem{choe2020attention}
Junsuk Choe, Seungho Lee, and Hyunjung Shim.
\newblock Attention-based dropout layer for weakly supervised single object
  localization and semantic segmentation.
\newblock {\em IEEE TPAMI}, 2020.

\bibitem{dai2015boxsup}
Jifeng Dai, Kaiming He, and Jian Sun.
\newblock Boxsup: Exploiting bounding boxes to supervise convolutional networks
  for semantic segmentation.
\newblock In {\em ICCV}, 2015.

\bibitem{everingham2010pascal}
Mark Everingham, Luc Van~Gool, Christopher~KI Williams, John Winn, and Andrew
  Zisserman.
\newblock The pascal visual object classes (voc) challenge.
\newblock {\em IJCV}, 88(2):303--338, 2010.

\bibitem{fan2020learning}
Junsong Fan, Zhaoxiang Zhang, Chunfeng Song, and Tieniu Tan.
\newblock Learning integral objects with intra-class discriminator for
  weakly-supervised semantic segmentation.
\newblock In {\em CVPR}, 2020.

\bibitem{fan2020cian}
Junsong Fan, Zhaoxiang Zhang, Tienniu Tan, Chunfeng Song, and Jun Xiao.
\newblock Cian: Cross-image affinity net for weakly supervised semantic
  segmentation.
\newblock In {\em AAAI}, 2020.

\bibitem{fu2019dual}
Jun Fu, Jing Liu, Haijie Tian, Yong Li, Yongjun Bao, Zhiwei Fang, and Hanqing
  Lu.
\newblock Dual attention network for scene segmentation.
\newblock In {\em CVPR}, 2019.

\bibitem{grill2020bootstrap}
Jean-Bastien Grill, Florian Strub, Florent Altch{\'e}, Corentin Tallec,
  Pierre~H Richemond, Elena Buchatskaya, Carl Doersch, Bernardo~Avila Pires,
  Zhaohan~Daniel Guo, Mohammad~Gheshlaghi Azar, et~al.
\newblock Bootstrap your own latent: A new approach to self-supervised
  learning.
\newblock In {\em NeurIPS}, 2020.

\bibitem{hariharan2011semantic}
Bharath Hariharan, Pablo Arbel{\'a}ez, Lubomir Bourdev, Subhransu Maji, and
  Jitendra Malik.
\newblock Semantic contours from inverse detectors.
\newblock In {\em ICCV}, 2011.

\bibitem{he2020momentum}
Kaiming He, Haoqi Fan, Yuxin Wu, Saining Xie, and Ross Girshick.
\newblock Momentum contrast for unsupervised visual representation learning.
\newblock In {\em CVPR}, 2020.

\bibitem{he2016deep}
Kaiming He, Xiangyu Zhang, Shaoqing Ren, and Jian Sun.
\newblock Deep residual learning for image recognition.
\newblock In {\em CVPR}, pages 770--778, 2016.

\bibitem{hou2018self}
Qibin Hou, PengTao Jiang, Yunchao Wei, and Ming-Ming Cheng.
\newblock Self-erasing network for integral object attention.
\newblock In {\em NeurIPS}, 2018.

\bibitem{huang2018weakly}
Zilong Huang, Xinggang Wang, Jiasi Wang, Wenyu Liu, and Jingdong Wang.
\newblock Weakly-supervised semantic segmentation network with deep seeded
  region growing.
\newblock In {\em CVPR}, 2018.

\bibitem{jiang2019integral}
Peng-Tao Jiang, Qibin Hou, Yang Cao, Ming-Ming Cheng, Yunchao Wei, and Hong-Kai
  Xiong.
\newblock Integral object mining via online attention accumulation.
\newblock In {\em ICCV}, 2019.

\bibitem{jin2021mining}
Zhenchao Jin, Tao Gong, Dongdong Yu, Qi Chu, Jian Wang, Changhu Wang, and Jie
  Shao.
\newblock Mining contextual information beyond image for semantic segmentation.
\newblock In {\em ICCV}, 2021.

\bibitem{ke2021universal}
Tsung-Wei Ke, Jyh-Jing Hwang, and Stella~X Yu.
\newblock Universal weakly supervised segmentation by pixel-to-segment
  contrastive learning.
\newblock In {\em ICLR}, 2021.

\bibitem{khoreva2017simple}
Anna Khoreva, Rodrigo Benenson, Jan Hosang, Matthias Hein, and Bernt Schiele.
\newblock Simple does it: Weakly supervised instance and semantic segmentation.
\newblock In {\em CVPR}, 2017.

\bibitem{khosla2020supervised}
Prannay Khosla, Piotr Teterwak, Chen Wang, Aaron Sarna, Yonglong Tian, Phillip
  Isola, Aaron Maschinot, Ce Liu, and Dilip Krishnan.
\newblock Supervised contrastive learning.
\newblock In {\em NeurIPS}, 2020.

\bibitem{kolesnikov2016seed}
Alexander Kolesnikov and Christoph~H Lampert.
\newblock Seed, expand and constrain: Three principles for weakly-supervised
  image segmentation.
\newblock In {\em ECCV}, 2016.

\bibitem{krahenbuhl2011efficient}
Philipp Kr{\"a}henb{\"u}hl and Vladlen Koltun.
\newblock Efficient inference in fully connected crfs with gaussian edge
  potentials.
\newblock {\em NeurIPS}, 2011.

\bibitem{Singh_2017}
Krishna Kumar~Singh and Yong Jae~Lee.
\newblock Hide-and-seek: Forcing a network to be meticulous for
  weakly-supervised object and action localization.
\newblock In {\em ICCV}, 2017.

\bibitem{lee2019ficklenet}
Jungbeom Lee, Eunji Kim, Sungmin Lee, Jangho Lee, and Sungroh Yoon.
\newblock Ficklenet: Weakly and semi-supervised semantic image segmentation
  using stochastic inference.
\newblock In {\em CVPR}, 2019.

\bibitem{lee2021bbam}
Jungbeom Lee, Jihun Yi, Chaehun Shin, and Sungroh Yoon.
\newblock Bbam: Bounding box attribution map for weakly supervised semantic and
  instance segmentation.
\newblock In {\em CVPR}, 2021.

\bibitem{lee2021railroad}
Seungho Lee, Minhyun Lee, Jongwuk Lee, and Hyunjung Shim.
\newblock Railroad is not a train: Saliency as pseudo-pixel supervision for
  weakly supervised semantic segmentation.
\newblock In {\em CVPR}, 2021.

\bibitem{li2018tell}
Kunpeng Li, Ziyan Wu, Kuan-Chuan Peng, Jan Ernst, and Yun Fu.
\newblock Tell me where to look: Guided attention inference network.
\newblock In {\em CVPR}, 2018.

\bibitem{li2021group}
Xueyi Li, Tianfei Zhou, Jianwu Li, Yi Zhou, and Zhaoxiang Zhang.
\newblock Group-wise semantic mining for weakly supervised semantic
  segmentation.
\newblock In {\em AAAI}, 2021.

\bibitem{li2021pseudo}
Yi Li, Zhanghui Kuang, Liyang Liu, Yimin Chen, and Wayne Zhang.
\newblock Pseudo-mask matters inweakly-supervised semantic segmentation.
\newblock In {\em ICCV}, 2021.

\bibitem{Liang2022TEL}
Zhiyuan Liang, Tiancai Wang, Xiangyu Zhang, Jian Sun, and Jianbing Shen.
\newblock Tree energy loss: Towards sparsely annotated semantic segmentation.
\newblock In {\em CVPR}, 2022.

\bibitem{lin2016scribblesup}
Di Lin, Jifeng Dai, Jiaya Jia, Kaiming He, and Jian Sun.
\newblock Scribblesup: Scribble-supervised convolutional networks for semantic
  segmentation.
\newblock In {\em CVPR}, 2016.

\bibitem{lin2014microsoft}
Tsung-Yi Lin, Michael Maire, Serge Belongie, James Hays, Pietro Perona, Deva
  Ramanan, Piotr Doll{\'a}r, and C~Lawrence Zitnick.
\newblock Microsoft coco: Common objects in context.
\newblock In {\em ECCV}, 2014.

\bibitem{liu2019simple}
Jiang-Jiang Liu, Qibin Hou, Ming-Ming Cheng, Jiashi Feng, and Jianmin Jiang.
\newblock A simple pooling-based design for real-time salient object detection.
\newblock In {\em CVPR}, 2019.

\bibitem{misra2020self}
Ishan Misra and Laurens van~der Maaten.
\newblock Self-supervised learning of pretext-invariant representations.
\newblock In {\em CVPR}, 2020.

\bibitem{oh2021background}
Youngmin Oh, Beomjun Kim, and Bumsub Ham.
\newblock Background-aware pooling and noise-aware loss for weakly-supervised
  semantic segmentation.
\newblock In {\em CVPR}, 2021.

\bibitem{oord2018representation}
Aaron van~den Oord, Yazhe Li, and Oriol Vinyals.
\newblock Representation learning with contrastive predictive coding.
\newblock {\em arXiv preprint arXiv:1807.03748}, 2018.

\bibitem{saleh2016built}
Fatemehsadat Saleh, Mohammad~Sadegh Aliakbarian, Mathieu Salzmann, Lars
  Petersson, Stephen Gould, and Jose~M Alvarez.
\newblock Built-in foreground/background prior for weakly-supervised semantic
  segmentation.
\newblock In {\em ECCV}, 2016.

\bibitem{shimoda2019self}
Wataru Shimoda and Keiji Yanai.
\newblock Self-supervised difference detection for weakly-supervised semantic
  segmentation.
\newblock In {\em ICCV}, 2019.

\bibitem{siam2019amp}
Mennatullah Siam, Boris~N Oreshkin, and Martin Jagersand.
\newblock Amp: Adaptive masked proxies for few-shot segmentation.
\newblock In {\em ICCV}, 2019.

\bibitem{simonyan2014very}
Karen Simonyan and Andrew Zisserman.
\newblock Very deep convolutional networks for large-scale image recognition.
\newblock In {\em ICLR}, 2015.

\bibitem{sohn2016improved}
Kihyuk Sohn.
\newblock Improved deep metric learning with multi-class n-pair loss objective.
\newblock In {\em NeurIPS}, 2016.

\bibitem{song2019box}
Chunfeng Song, Yan Huang, Wanli Ouyang, and Liang Wang.
\newblock Box-driven class-wise region masking and filling rate guided loss for
  weakly supervised semantic segmentation.
\newblock In {\em CVPR}, 2019.

\bibitem{sun2020mining}
Guolei Sun, Wenguan Wang, Jifeng Dai, and Luc Van~Gool.
\newblock Mining cross-image semantics for weakly supervised semantic
  segmentation.
\newblock In {\em ECCV}, 2020.

\bibitem{tian2020contrastive}
Yonglong Tian, Dilip Krishnan, and Phillip Isola.
\newblock Contrastive multiview coding.
\newblock In {\em ECCV}, 2020.

\bibitem{vernaza2017learning}
Paul Vernaza and Manmohan Chandraker.
\newblock Learning random-walk label propagation for weakly-supervised semantic
  segmentation.
\newblock In {\em CVPR}, 2017.

\bibitem{wang2021multiple}
Binglu Wang, Yongqiang Zhao, and Xuelong Li.
\newblock Multiple instance graph learning for weakly supervised remote sensing
  object detection.
\newblock {\em IEEE TGRS}, 60:1--12, 2021.

\bibitem{wang2021survey}
Wenguan Wang, Tianfei Zhou, Fatih Porikli, David Crandall, and Luc Van~Gool.
\newblock A survey on deep learning technique for video segmentation.
\newblock {\em arXiv preprint arXiv:2107.01153}, 2021.

\bibitem{wang2021hierarchical}
Wenguan Wang, Tianfei Zhou, Siyuan Qi, Jianbing Shen, and Song-Chun Zhu.
\newblock Hierarchical human semantic parsing with comprehensive part-relation
  modeling.
\newblock {\em IEEE TPAMI}, 2021.

\bibitem{wang2021exploring}
Wenguan Wang, Tianfei Zhou, Fisher Yu, Jifeng Dai, Ender Konukoglu, and Luc
  Van~Gool.
\newblock Exploring cross-image pixel contrast for semantic segmentation.
\newblock In {\em ICCV}, 2021.

\bibitem{wang2020weakly}
Xiang Wang, Sifei Liu, Huimin Ma, and Ming-Hsuan Yang.
\newblock Weakly-supervised semantic segmentation by iterative affinity
  learning.
\newblock {\em IJCV}, pages 1--14, 2020.

\bibitem{wang2018weakly}
Xiang Wang, Shaodi You, Xi Li, and Huimin Ma.
\newblock Weakly-supervised semantic segmentation by iteratively mining common
  object features.
\newblock In {\em CVPR}, 2018.

\bibitem{wang2020cross}
Xun Wang, Haozhi Zhang, Weilin Huang, and Matthew~R Scott.
\newblock Cross-batch memory for embedding learning.
\newblock In {\em CVPR}, 2020.

\bibitem{wang2021dense}
Xinlong Wang, Rufeng Zhang, Chunhua Shen, Tao Kong, and Lei Li.
\newblock Dense contrastive learning for self-supervised visual pre-training.
\newblock In {\em CVPR}, 2021.

\bibitem{wang2020self}
Yude Wang, Jie Zhang, Meina Kan, Shiguang Shan, and Xilin Chen.
\newblock Self-supervised equivariant attention mechanism for weakly supervised
  semantic segmentation.
\newblock In {\em CVPR}, 2020.

\bibitem{wei2017object}
Yunchao Wei, Jiashi Feng, Xiaodan Liang, Ming-Ming Cheng, Yao Zhao, and
  Shuicheng Yan.
\newblock Object region mining with adversarial erasing: A simple
  classification to semantic segmentation approach.
\newblock In {\em CVPR}, 2017.

\bibitem{wei2016stc}
Yunchao Wei, Xiaodan Liang, Yunpeng Chen, Xiaohui Shen, Ming-Ming Cheng, Jiashi
  Feng, Yao Zhao, and Shuicheng Yan.
\newblock Stc: A simple to complex framework for weakly-supervised semantic
  segmentation.
\newblock {\em IEEE TPAMI}, 39(11):2314--2320, 2016.

\bibitem{wei2018revisiting}
Yunchao Wei, Huaxin Xiao, Honghui Shi, Zequn Jie, Jiashi Feng, and Thomas~S
  Huang.
\newblock Revisiting dilated convolution: A simple approach for weakly-and
  semi-supervised semantic segmentation.
\newblock In {\em CVPR}, 2018.

\bibitem{wu2021embedded}
Tong Wu, Junshi Huang, Guangyu Gao, Xiaoming Wei, Xiaolin Wei, Xuan Luo, and
  Chi~Harold Liu.
\newblock Embedded discriminative attention mechanism for weakly supervised
  semantic segmentation.
\newblock In {\em CVPR}, 2021.

\bibitem{wu2018unsupervised}
Zhirong Wu, Yuanjun Xiong, Stella~X Yu, and Dahua Lin.
\newblock Unsupervised feature learning via non-parametric instance
  discrimination.
\newblock In {\em CVPR}, 2018.

\bibitem{xiao2017joint}
Tong Xiao, Shuang Li, Bochao Wang, Liang Lin, and Xiaogang Wang.
\newblock Joint detection and identification feature learning for person
  search.
\newblock In {\em CVPR}, 2017.

\bibitem{xie2021propagate}
Zhenda Xie, Yutong Lin, Zheng Zhang, Yue Cao, Stephen Lin, and Han Hu.
\newblock Propagate yourself: Exploring pixel-level consistency for
  unsupervised visual representation learning.
\newblock In {\em CVPR}, 2021.

\bibitem{xu2021leveraging}
Lian Xu, Wanli Ouyang, Mohammed Bennamoun, Farid Boussaid, Ferdous Sohel, and
  Dan Xu.
\newblock Leveraging auxiliary tasks with affinity learning for weakly
  supervised semantic segmentation.
\newblock In {\em ICCV}, 2021.

\bibitem{yao2021non}
Yazhou Yao, Tao Chen, Guo-Sen Xie, Chuanyi Zhang, Fumin Shen, Qi Wu, Zhenmin
  Tang, and Jian Zhang.
\newblock Non-salient region object mining for weakly supervised semantic
  segmentation.
\newblock In {\em CVPR}, 2021.

\bibitem{yuan2020object}
Yuhui Yuan, Xilin Chen, and Jingdong Wang.
\newblock Object-contextual representations for semantic segmentation.
\newblock In {\em ECCV}, 2020.

\bibitem{zeng2019joint}
Yu Zeng, Yunzhi Zhuge, Huchuan Lu, and Lihe Zhang.
\newblock Joint learning of saliency detection and weakly supervised semantic
  segmentation.
\newblock In {\em ICCV}, 2019.

\bibitem{zhang2020causal}
Dong Zhang, Hanwang Zhang, Jinhui Tang, Xiansheng Hua, and Qianru Sun.
\newblock Causal intervention for weakly-supervised semantic segmentation.
\newblock In {\em NeurIPS}, 2020.

\bibitem{zhang2021complementary}
Fei Zhang, Chaochen Gu, Chenyue Zhang, and Yuchao Dai.
\newblock Complementary patch for weakly supervised semantic segmentation.
\newblock In {\em ICCV}, 2021.

\bibitem{zhang2018mixup}
Hongyi Zhang, Moustapha Cisse, Yann~N Dauphin, and David Lopez-Paz.
\newblock mixup: Beyond empirical risk minimization.
\newblock In {\em ICLR}, 2018.

\bibitem{zhang2018context}
Hang Zhang, Kristin Dana, Jianping Shi, Zhongyue Zhang, Xiaogang Wang, Ambrish
  Tyagi, and Amit Agrawal.
\newblock Context encoding for semantic segmentation.
\newblock In {\em CVPR}, 2018.

\bibitem{Zhang_2018}
Xiaolin Zhang, Yunchao Wei, Jiashi Feng, Yi Yang, and Thomas~S. Huang.
\newblock Adversarial complementary learning for weakly supervised object
  localization.
\newblock In {\em CVPR}, 2018.

\bibitem{zhang2020inter}
Xiaolin Zhang, Yunchao Wei, and Yi Yang.
\newblock Inter-image communication for weakly supervised localization.
\newblock In {\em ECCV}, 2020.

\bibitem{zhou2016learning}
Bolei Zhou, Aditya Khosla, Agata Lapedriza, Aude Oliva, and Antonio Torralba.
\newblock Learning deep features for discriminative localization.
\newblock In {\em CVPR}, 2016.

\bibitem{zhou2021target}
Tianfei Zhou, Jianwu Li, Xueyi Li, and Ling Shao.
\newblock Target-aware object discovery and association for unsupervised video
  multi-object segmentation.
\newblock In {\em CVPR}, 2021.

\bibitem{zhou2021group}
Tianfei Zhou, Liulei Li, Xueyi Li, Chun-Mei Feng, Jianwu Li, and Ling Shao.
\newblock Group-wise learning for weakly supervised semantic segmentation.
\newblock {\em IEEE TIP}, 31:799--811, 2021.

\bibitem{zhou2020motion}
Tianfei Zhou, Shunzhou Wang, Yi Zhou, Yazhou Yao, Jianwu Li, and Ling Shao.
\newblock Motion-attentive transition for zero-shot video object segmentation.
\newblock In {\em AAAI}, 2020.

\bibitem{zhou2021differentiable}
Tianfei Zhou, Wenguan Wang, Si Liu, Yi Yang, and Luc Van~Gool.
\newblock Differentiable multi-granularity human representation learning for
  instance-aware human semantic parsing.
\newblock In {\em CVPR}, 2021.

\end{thebibliography}
}

\end{document}